\pdfoutput=1
\documentclass[letterpaper]{article} 
\usepackage{aaai23}  
\usepackage{times}  
\usepackage{helvet}  
\usepackage{courier}  
\usepackage[hyphens]{url}  
\usepackage{graphicx} 
\usepackage{natbib}  
\usepackage{caption} 
\usepackage{amsmath,amssymb} 
\usepackage{algorithm}
\usepackage{algorithmic}
\usepackage{makecell}
\usepackage{multirow}
\usepackage{booktabs}

\usepackage{newfloat}
\usepackage{listings}
\DeclareCaptionStyle{ruled}{labelfont=normalfont,labelsep=colon,strut=off} 
\floatstyle{ruled}
\newfloat{listing}{tb}{lst}{}
\floatname{listing}{Listing}
\title{Improving Crowded Object Detection via Copy-Paste}
\author{
    Jiangfan Deng,
    Dewen Fan,
    Xiaosong Qiu,
    Feng Zhou
}
\affiliations{
    Algorithm Research, Aibee Inc.\\

    jfdeng100@foxmail.com,
    \{dwfan,xsqiu,fzhou\}@aibee.com
}

\usepackage{bibentry}

\begin{document}

\maketitle

\begin{abstract}
Crowdedness caused by overlapping among similar objects is a ubiquitous challenge in the field of 2D visual object detection.
In this paper, we first underline two main effects of the crowdedness issue:
1) IoU-confidence correlation disturbances (ICD) and 2) confused de-duplication (CDD).
Then we explore a pathway of cracking these nuts from the perspective of data augmentation.
Primarily, a particular copy-paste scheme is proposed towards making crowded scenes.
Based on this operation, we first design a ``consensus learning'' strategy to further resist the ICD problem
and then find out the pasting process naturally reveals a pseudo ``depth'' of object in the scene,
which can be potentially used for alleviating CDD dilemma.
Both methods are derived from magical using of the copy-pasting without extra cost for hand-labeling.
Experiments show that our approach can easily improve the state-of-the-art detector in typical crowded detection task by more than 2\% without any bells and whistles.
Moreover, this work can outperform existing data augmentation strategies in crowded scenario.
\end{abstract}

\section{Introduction}
\label{sec:intro}

\begin{figure}[t]
  \centering
  \includegraphics[width=1.\linewidth]{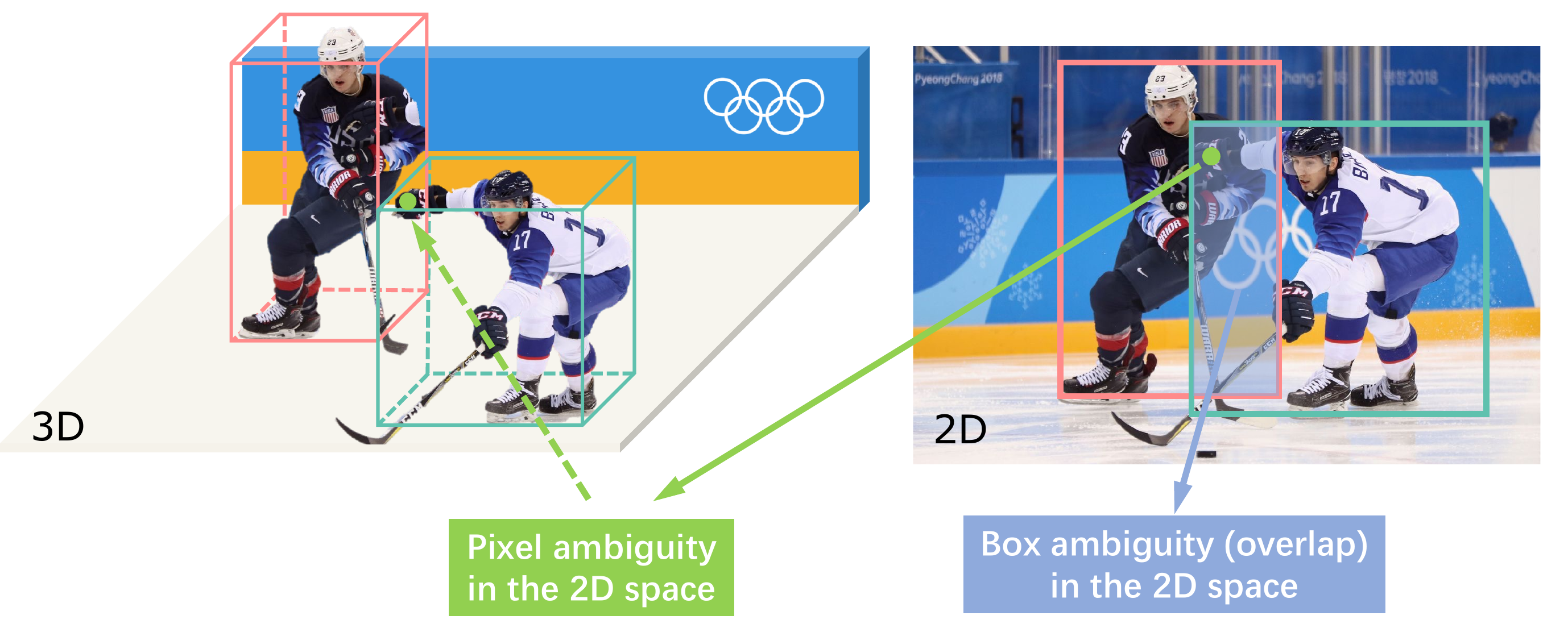}
  \vspace{-1.5\baselineskip}
  \caption{{\bf{Semantic ambiguities in the 2D space.}}
    We exhibit the same scenario in the real 3D world (left) and the 2D space after photographing (right) respectively.
    The colored boxes represent two distinct objects (pucksters) while the \textit{green} points denote a voxel in 3D space and its corresponding pixel in the 2D image.
    It is clearly illustrated that the 3D voxel lies on the body of a unique puckster while the 2D pixel lies on both of them.
    After evolving from a point to a bounding-box, the ambiguity arises in the form of overlap. 
            }
\label{fig:2d-ambiguity}
\vspace{-0.3cm}
\end{figure}

The task of object detection has been meticulously studied for quite a long time.
In the deep learning era, in recent years, many well-designed methods~\cite{LiuOWFCLP20} have been proposed and raised the detection performance to a surprisingly high level.
Nevertheless, there still exist many intrinsic problems that are not fundamentally solved.
One of them is the ``crowdedness issue'', which usually denotes the phenomenon that objects belonging to the same category are highly overlapped together.
In a geometrical manner, the basic difficulty stems from the semantical ambiguities of the 2D space.
As shown in Fig.~\ref{fig:2d-ambiguity}, in our 3D world, each voxel has its ``unique semantics'' and lies on a ``certain object''.
However, after projecting to 2D plane, one pixel might fall on several collided objects.
After evolving the concept from a ``pixel'' to a ``box'', the semantical ambiguity in crowded scenes leads to the notion of \emph{overlap}.
%
%

\begin{figure}[t]
  \centering
  \includegraphics[width=1.\linewidth]{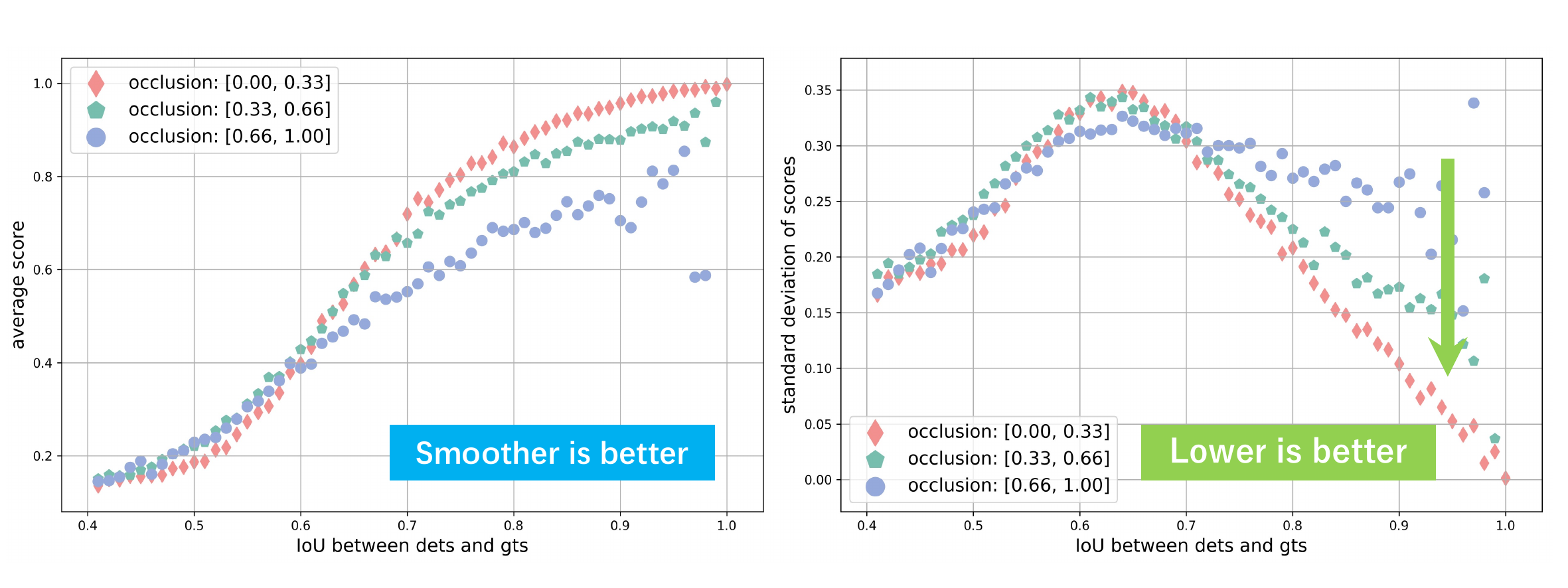}
  \vspace{-1.5\baselineskip}
  \caption{ \textbf{IoU-confidence correlation disturbances (ICD).}
            We visualize the confidence score w.r.t the IoU between the predicted box from \cite{he2017mask} and ground-truth in CrowdHuman~\cite{shao2018crowdhuman}.
            First, the IoU range of $[0, 1]$ are equally divided into 100 bins (each with the length of 0.01) as the horizontal axis.
            Then, \emph{average value} (left) or \emph{standard deviation} (right) of confidence scores are computed within each bin, generating a corresponding point in the coordinate plane.
            Marker shapes of diamond (\textit{red}), pentagon (\textit{green}) and circle (\textit{blue}) refer to crowdedness degrees with the occlusion ratio on three ranges of $[0, 0.33]$, $[0.33, 0.66]$ and $[0.66, 1]$ respectively.
            On the left figure, the average score curve corresponding to the most crowded range (blue) are obviously \textbf{more jittering} than the other two curves;
            On the right figure, the heavier the crowdedness is, the \textbf{larger} the standard deviations are.
            Both figures suggest that the IoU-confidence correlation would become more \textbf{uncertain} when the crowdedness increases.
            }
\label{fig:icd-issue}
\end{figure}

To probe the effects of this problem, we now dive into the essence of the detection paradigm. 
Generally, an object detector reads in an image and outputs a set of bounding-boxes each associated with a confidence score.
For an ideally-performed detector, the score value should convey how well the predicted box is overlapped with the ground-truth. In other words, the Intersection-over-Union (IoU) between these two boxes should be positively correlated with the confidence score.
After visualizing the mean and standard deviation of scores with respect to IoU in Fig.~\ref{fig:icd-issue}, it turns out that even for the off-the-shelf detectors like~\cite{he2017mask}, this positive correlation would be gradually disturbed by the increase of crowdedness degree\footnote{The crowdedness degree is indicated in terms of ``occlusion ratio'', \emph{i.e.}, $1 - s_v/s_f$, where the $s_v$ and $s_f$ represent size of the visible box and full box of an object.}.
This experimental study clearly indicates the struggle of current detection algorithms in facing the super-heavy overlaps. 
We embody this effect as IoU-confidence Correlation Disturbances (ICD).
On the other hand, a typical detection pipeline often ends with a de-duplication module,
for example, the widely adopted Non-Maximum Suppression (NMS).
Due to the 2D semantical ambiguity mentioned previously, these modules are often confused by heavily overlapped predictions,
which leads to severe missing in a crowd.
We cast this type of effect as Confused De-Duplication (CDD).
%
%
%

To overcome these two obstacles, we explore a pathway from the perspective of data augmentation.
Referring to the preceding works~\cite{DBLP:conf/cvpr/GhiasiCSQLCLZ21,DBLP:conf/iccv/DwibediMH17,DBLP:conf/cvpr/LiSYP21,dvornik2018modeling,DBLP:conf/iccv/FangSWGLL19}, a simple copy-paste variant is proposed.
Firstly, object segmentation patches are pasted to the training images following some specialized rules dedicated for making crowded scenes.
Then, revolved from copy-pasting, we design a ``consensus learning'' approach to align confidence distributions of overlaid objects to their \emph{identical but non-overlaid} counterparts, which further restrains the ICD problem.
Moreover, thanks to the program-controlled pasting process,
we can naturally get the extra order information of which one is in the front and which one is in the back when two (pasted) objects are overlapped.
This cost-free knowledge provides cues on the additional \emph{third dimension of depth} apart from \emph{x} and \emph{y}-axis spanning the image plane,
which can be deemed as a breakthrough of the aforementioned 2D restrictions inducing the CDD dilemma.
From this motivation, we propose a concept named ``overlay depth'' and semi-supervisely train the detector to predict this label. 
Then, an Overlay Depth-aware NMS (OD-NMS) is introduced to make use of the depth knowledge during de-duplication.
Experiments show that this strategy can help distinguish boxes gathered in 2D space and further boost the detection results.

We evaluate our method from multiple aspects.
As a data augmentation strategy, this work can outperform other counterparts in crowded scenes, no matter hand-craft methods or automated ones.
As an approach of countering crowdedness issue, our method can stably improve the state-of-the-art detector by more than 2\% without any bells and whistles.
Moreover, since hand-labeling the crowded data is resource-consuming,
this method provides a way of training on ``sparse data'' only and applying to crowded scenes via data augmentation.

To sum up, the major contributions of this work are two-fold:
(1) We propose a crowdedness-oriented copy-paste scheme and introduce a consensus learning strategy,
which effectively helps the detector resisting the ICD problem and bring improvements in crowded scenes.
(2) We design a simple method to utilize the weak depth knowledge produced by the pasting process, which further optimize the detector.

\section{Related Works}
\label{sec:related}
{\noindent\bf{Crowded Object Detection.}}
Detecting objects in crowded scenes has been a long-standing challenge~\cite{LiuOWFCLP20} and much effort has been spent on this topic.
For example, \cite{wang2018repulsion} and~\cite{zhang2018occlusion} propose specific loss functions to constrain proposals closer to the corresponding ground-truth and further away from the nearby objects, thereby enhancing discrimination between overlapped individuals.
CaSe~\cite{xie2020count} uses a new branch to count pedestrian number in a region of interest (RoI) and generates similarity embeddings for each proposal.
As a response to the CDD problem mentioned above, a group of works focuses on alleviating the deficiency of Non-Maximum Suppression (NMS). 
Adaptive-NMS~\cite{liu2019adaptive} introduces an adaptation mechanism to dynamically adjust the threshold in NMS, leading to better recall in a crowd.
In~\cite{gahlert2020visibility} and~\cite{huang2020visible}, NMS leverages the less-occluded visible boxes to guide the selection of full boxes,
whereas extra labeling (of the visible boxes) is required.
CrowdDet~\cite{chu2020detection} conducts one proposal to make multiple predictions and uses an artfully designed Set-NMS to solve heavily-overlapped cases.
Some recent works explore other ways.
~\cite{zhang2021variational} models the pedestrian detection task as a variational inference problem.
~\cite{zheng2022progressive} refines the end-to-end detector Sparse R-CNN~\cite{sun2021sparse} to adapt to the crowded detection scenario.
\\[5pt]
{\bf{Data Augmentation in Object Detection.}}
In the field of computer vision, data augmentation~\cite{ShortenK19} has long been used to optimize the model training,
which originates mainly from the image classification task~\cite{he2016deep,tan2019efficientnet}.
Early approaches usually include strategies such as color shifting~\cite{2014Going} and random crop~\cite{DBLP:conf/nips/KrizhevskySH12,DBLP:journals/pieee/LeCunBBH98,DBLP:journals/corr/SimonyanZ14a,2014Going}.
Naturally, the core ideas were transferred to the detection domain and some operations (\emph{e.g.}, image flipping and
scale jittering) have been widely adopted as a standard module~\cite{liu2016ssd,redmon2016you,Ren2015Faster}. 
Currently, methods with more concrete theoretical basis have emerged.
These variants, ranging from hand-crafted Cutout~\cite{DBLP:journals/corr/abs-1708-04552}, Mixup~\cite{2017mixup} and CutMix~\cite{0CutMix} to learning based AutoAugment~\cite{2018AutoAugment}, Fast AutoAugment~\cite{DBLP:conf/nips/LimKKKK19} and RandAugment~\cite{DBLP:conf/nips/CubukZS020}, perform considerable effects on image classification and suggest huge potential in object detection.
Meanwhile, there are also some works focusing on detection task.
Stitcher~\cite{DBLP:journals/corr/abs-2004-12432} and YOLOv4~\cite{DBLP:journals/corr/abs-2004-10934} introduce mosaic inputs containing rescaled image patches to enhance robustness.
~\cite{DBLP:conf/eccv/ZophCGLSL20} and ~\cite{chen2021scale} re-design the AutoAugment scheme to adapt to object detection.
In~\cite{tang2021autopedestrian}, researchers propose a method searching the policy of data augmentation and loss function jointly.
In~\cite{liu2020novel}, a novel APGAN is proposed to transfer pedestrians from other datasets in making augmentation.
\\[5pt]
{\bf{Copy-Paste Augmentation.}} Copy-paste augmentation is first invented in~\cite{DBLP:conf/iccv/DwibediMH17}.
By cutting object patches from the source image and pasting to the target one,
a combinatorial amount of synthetic training data can be easily acquired and improve the detection/segmentation performance significantly.
This amazing magic power is then verified by subsequent works~\cite{DBLP:conf/eccv/RemezHB18,DBLP:conf/cvpr/LiSYP21,DBLP:conf/iccv/FangSWGLL19,dvornik2018modeling,DBLP:conf/cvpr/GhiasiCSQLCLZ21} and the method has been further polished by context adaptation~\cite{DBLP:conf/iccv/FangSWGLL19,DBLP:conf/eccv/RemezHB18,dvornik2018modeling}.
In~\cite{DBLP:conf/cvpr/GhiasiCSQLCLZ21}, the authors claim that simple copy-paste can bring considerable improvement as long as the training is sufficient enough.
Their experiments further suggest the potential of this augmentation strategy on instance-level image understanding.
It should be noted that the initial motivation of copy-paste is to diversify the sample space, especially for the rare categories~\cite{DBLP:conf/cvpr/GhiasiCSQLCLZ21} or alleviating the complex mask labeling~\cite{DBLP:conf/eccv/RemezHB18}.
However, in our work, we utilize this operation to precisely solve the crowdedness issue.
Although there has been simple practice in previous works~\cite{DBLP:conf/iccv/DwibediMH17,DBLP:conf/cvpr/GhiasiCSQLCLZ21},
the actual effects of this strategy on dealing with crowdedness scenario has never been systematically designed and studied.

\section{Resist the IoU-Confidence Disturbances}
\label{sec:ccp}
This part focuses on solving the Iou-Confidence Disturbances (ICD).
We explore two consecutive ways in achieving this aim.
First, doing copy-paste to make crowded scenes. 
Then, introducing consensus learning between overlaid objects and their non-overlaid counterparts,
which relies on the copy-pasting.

\subsection{Crowdedness Oriented Copy-Paste}
Based on observations of Fig.~\ref{fig:icd-issue},
an intuitive idea is to make more crowded cases to dominate the training.
To this end, we carefully re-design the copy-paste strategy.
First, the conception of ``group'' is introduced. 
An image should include several groups and each group consists of multiple heavily overlapped objects.
Following this logic scheme, we first generate the group centers on an image and then paste objects around them.

Formally, for every training image to be augmented, we initialize a set $\mathcal{C}$ of ``group centers'':
\begin{equation}\notag
\label{eq:crowd_ctrs}
    \mathcal{C} = \{(x_1, y_1, s_1),..., (x_{|\mathcal{C}|}, y_{|\mathcal{C}|}, s_{|\mathcal{C}|})\},
\end{equation}
where each tuple represents the object locating at center of the corresponding group
($x_i$, $y_i$ and $s_i$ denote the coordinates and normalized object size respectively).
We obtain these group centers by sampling from original objects on the current image.
The group number $|\mathcal{C}|$ is randomly chosen from an integral range of $[0, N]$, where $\emph{N}$ is a hyper parameter.

The second step is pasting objects around these group centers.
For each $c_i \in \mathcal{C}$, we should generate a set $\hat{\mathcal{G}}_i$ of objects in the group $i$:
\begin{equation}\notag
\label{eq:crowd_objs}
    \hat{\mathcal{G}}_i = \{(\hat{x}_1^{i}, \hat{y}_1^{i}, \hat{s}_1^{i}),..., (\hat{x}_{|\hat{\mathcal{G}}_i|}^{i}, \hat{y}_{|\hat{\mathcal{G}}_i|}^{i}, \hat{s}_{|\hat{\mathcal{G}}_i|}^{i})\},
\end{equation}
similarly, object number $|\hat{\mathcal{G}}_i|$ in the group comes from range $[0, M]$ where $\emph{M}$ is another hyper parameter.
Since the nature of crowdedness is ``overlapping'',
every $\hat{g}_j^i \in \hat{\mathcal{G}}_i$ is enforced to be overlapped with the group center object $c_i$.
We manipulate the overlapping from three aspects of the $\emph{x}$, $\emph{y}$ and $\emph{s}$ conditioning in a probabilistic sense.

First, objects in a group usually have similar sizes.
Let $p(\hat{s}_j^i | s_i, I)$ be the probability density function of $\hat{s}_j^i$ on conditions of the center object size $s_i$ in the image $\emph{I}$.
We choose $p(\cdot)$ to be a Gaussian as:
\begin{equation}
\label{eq:gaussian_sji}
    p(\hat{s}_j^i | s_i, I) = \frac{1}{\sqrt{2\pi}\sigma} \exp({-\frac{(\hat{s}_j^i - s_i)^2}{2\sigma^2}}),
\end{equation}
where $\sigma$ is the standard deviation which a constant value 0.2 is used in this paper.
To guarantee overlapping, we adopt two independent uniform distributions in modeling
the coordinate values $\hat{x}_j^{i}$ and $\hat{y}_j^{i}$:
\begin{align}
\label{eq:uniform_xy}
    \hat{x}_j^i \sim U(x_i-\frac{d_w}{\tau}, x_i+\frac{d_w}{\tau}),\\
    \hat{y}_j^i \sim U(y_i-\frac{d_h}{\epsilon}, y_i+\frac{d_h}{\epsilon}),
\end{align}
where $d_w$ and $d_h$ are the maximum distances of $\hat{g}_j^i$ shifting from group center $c_i$ with overlap.
Coefficients $\tau > 1$ and $\epsilon > 1$ are used to adjust the crowdedness degree.

During training, for every image loaded, the set $\mathcal{C}$ and $\hat{\mathcal{G}}_i$-s are generated obeying rules above.
Then object segmentation patches would be sampled, re-scaled and pasted to the image accordingly.

\subsection{Consensus Learning}
\begin{figure}[t]
  \centering
  \includegraphics[width=1.\linewidth]{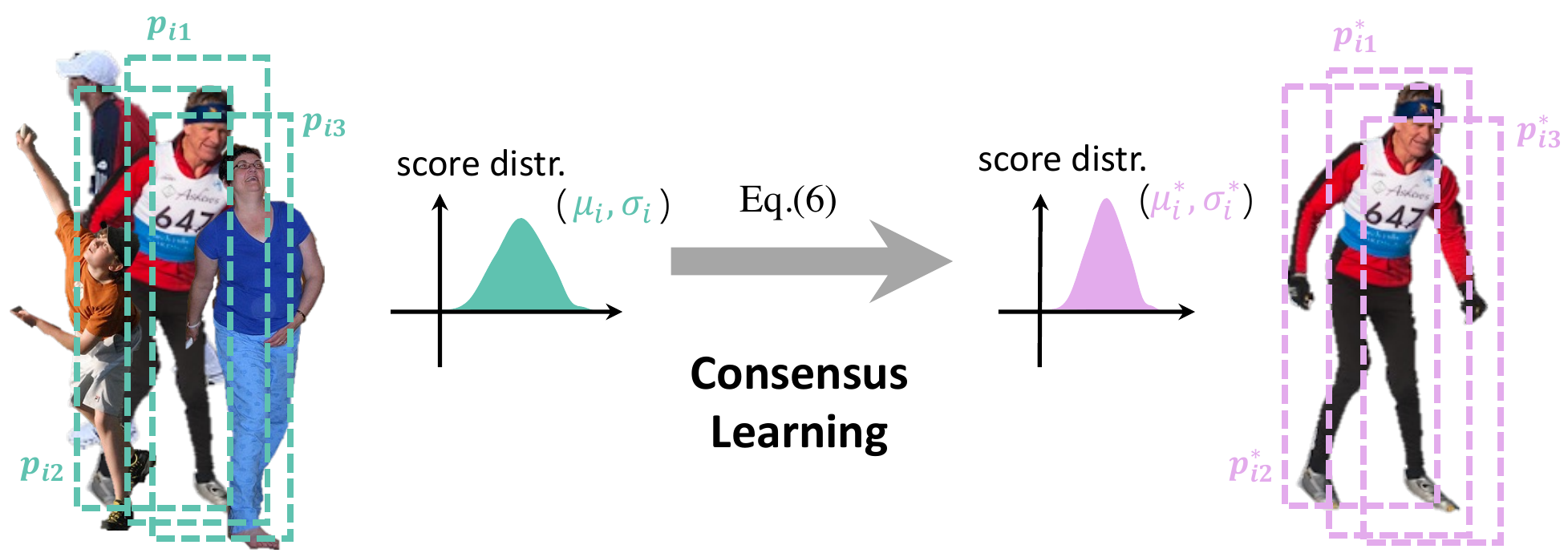}
  \vspace{-1.5\baselineskip}
  \caption{{\textbf{Consensus Learning.}
      Learn to reach consensus between the overlaid object (the man in red on the left) and its identical but non-overlaid counterpart (right).}
  }
\label{fig:cl}
\end{figure}

With the toolkit of copy-pasting, we augment detector training with a dedicated strategy for resisting the ICD issue.
Given the observation shown in Fig.~\ref{fig:icd-issue} that the instability of predicted scores derives from crowdedness,
an emerging fix is to align the score of an object in crowded circumstances (overlaid by other objects) to that when it is not overlaid.
Thanks to the copy-paste method, we can easily generate this type of object pairs in which two identical objects lie in different surroundings.
Fig.~\ref{fig:cl} illustrates our idea.
Following the previous data augmentation, we pick out a set $\mathcal{B}_{ovl}$ of objects which are overlaid by others.
Then, the same object patches with those in $\mathcal{B}_{ovl}$ are re-pasted to the image without been overlaid,
constructing another set $\mathcal{B}^{*}_{ovl}$.
During training, we enforce the predicted score distributions of each object $b_{i}\in\mathcal{B}_{ovl}$
in an alignment with its counterpart $b^{*}_{i}\in\mathcal{B}^{*}_{ovl}$.
We term this process as \emph{consensus learning} by drawing an analogy of ``reaching consensus'' within each pair.
Specifically, let $\mathcal{P}_i$ be the set of proposals matched to $b_i$ and $\mathcal{P}^{*}_i$ be the set of proposals matched to $b^{*}_i$.
We first compute the mean $\mu$ and standard deviation $\sigma$ of scores for each object:
\small
\begin{equation}
\mu_i=\frac{1}{m}\sum_{p_{ij}\in\mathcal{P}_i}c(p_{ij}),\quad\sigma_i=\sqrt{\frac{1}{m}\sum_{p_{ij}\in\mathcal{P}_i}(c(p_{ij})-\mu_i)^2},\\
\end{equation}
\begin{equation}
\mu^*_i=\frac{1}{m^*}\sum_{p^*_{ij}\in\mathcal{P}^*_i}c(p^*_{ij}),\enspace\sigma^*_i=\sqrt{\frac{1}{m^*}\sum_{p^*_{ij}\in\mathcal{P}^*_{ij}}(c(p^*_i)-\mu^*_i)^2},
\end{equation}
\normalsize
where $m$ and $m^*$ are the sizes of $\mathcal{P}_i$ and $\mathcal{P}^*_i$ respectively and $c(\cdot)$ denotes the predicted confidence score of a proposal.
Then we pursue a pair of $\{\mu_i, \sigma_i\}$ approaching $\{\mu^*_i, \sigma^*_i\}$ through the mean squared error (MSE) loss:
\begin{equation}
L_{cl} = \frac{1}{|\mathcal{B}_{ovl}|}\sum_{b_i\in\mathcal{B}_{ovl}}(\mu_i-\mu^*_i)^2 + (\sigma_i - \sigma^*_i)^2.
\end{equation}
It is worth to point that only the overlaid half $\{\mu_i, \sigma_i\}$ contributes to the gradient back-propagation
while the non-overlaid half (marked by $*$) is treated as target.
%
%

\subsection{Analyze the IoU-Confidence Disturbances}
\label{sec:resist-icd}
\begin{figure}[t]
  \centering
  \includegraphics[width=1.\linewidth]{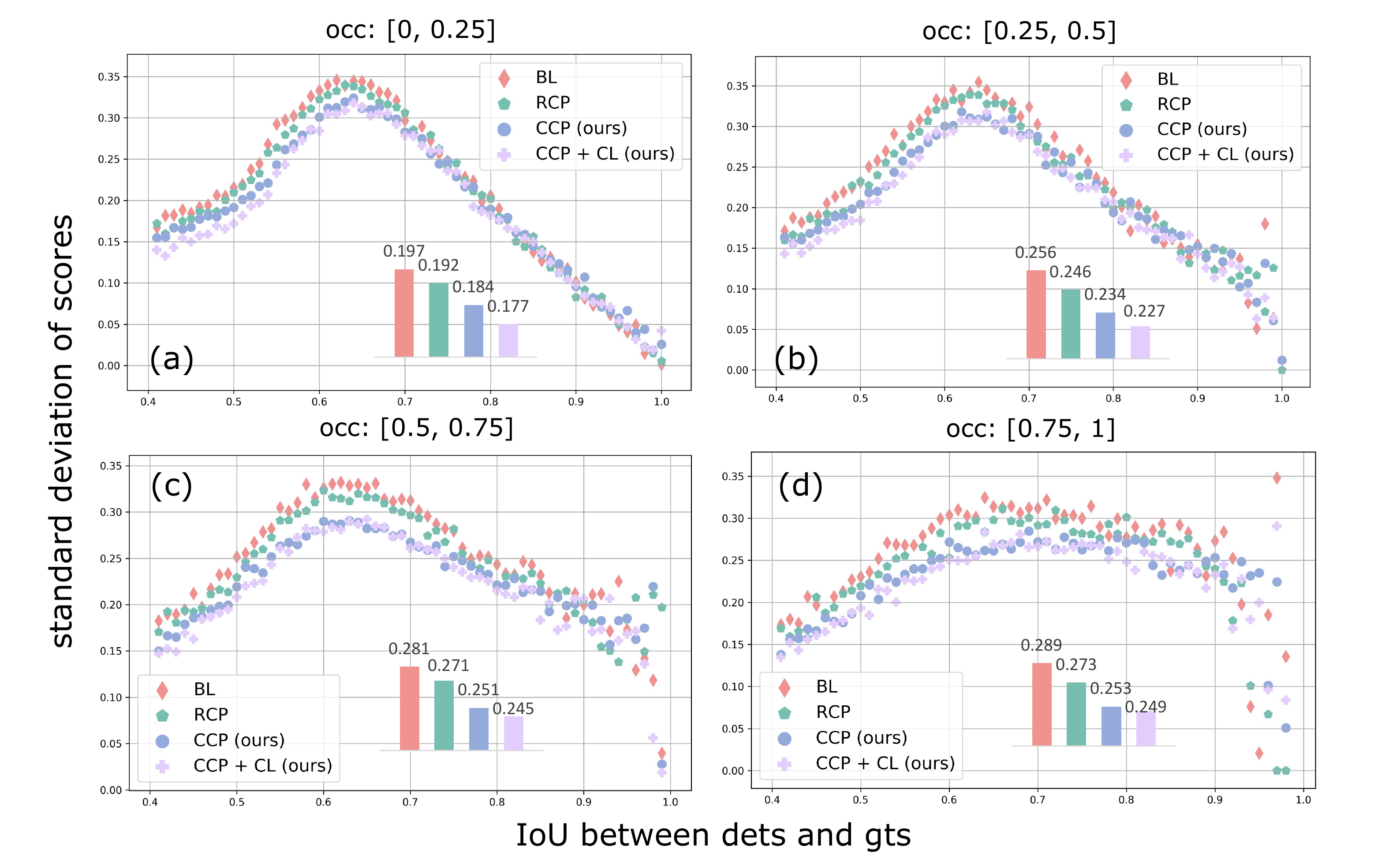}
  \vspace{-1.5\baselineskip}
  \caption{{\bf{Effects of our method on the ICD issue, lower is better.}}
    We plot only the \emph{standard deviation} of confidence scores w.r.t the IoU value on CrowdHuman.
    The crowdedness (occlusion ratio) gradually increases from (a) to (d).
  }
\label{fig:resist-icd}
\end{figure}

Now we analyze the effectiveness of our method on mitigating the aforementioned ICD issue.
To revisit the original motivation raised from the right of Fig.~\ref{fig:icd-issue}, we plot the standard deviation (STD) of scores in Fig.~\ref{fig:resist-icd}.
First, it is clearly demonstrated that score STDs of the model trained with our Crowdedness-oriented Copy-Paste (CCP) are obviously \emph{\textbf{lower}} than those of the baseline model (BL) and the gap becomes larger by improving the crowdedness degree (from Fig.~\ref{fig:resist-icd}-(a) to (d)).
Second, although the curves of CCP and CCP+CL seems with no clear distinction,
after computing their average STDs (the four histograms in Fig.~\ref{fig:resist-icd}), we find the value of the latter is actually lower than that of the former.
Moreover, we plot another model augmented with random copy-paste (RCP) without specially taking crowdedness into consideration.
It is obvious that the decline of score STDs is with a much smaller margin.
These observations convince that our method can significantly improve the detector's robustness in crowded scenes and therefore alleviate the ICD problem.
%
%

\section{Alleviate the Confused De-Duplications}
\label{sec:od}
Our augmentation strategy has a natural by-product: for these overlapped objects pasted, the relative ``order of depth'' is known a priori.
In other words, we are aware of which one is in the front and which one is in the back.
Now let us return to the semantical ambiguity described in Sec.\ref{sec:intro}.
Basically, ambiguities in 2D space are caused by the absence of one dimension in the real (3D) world.
From this point of view, the depth order can be viewed as some weak knowledge of the additional \emph{third dimension}, which shed light on mitigating the vagueness.
As a feasible practice, in this work, we utilize the depth order information to resolve the confused de-duplication (CDD) problem.

\begin{figure}[t]
  \centering
  \includegraphics[width=1.\linewidth]{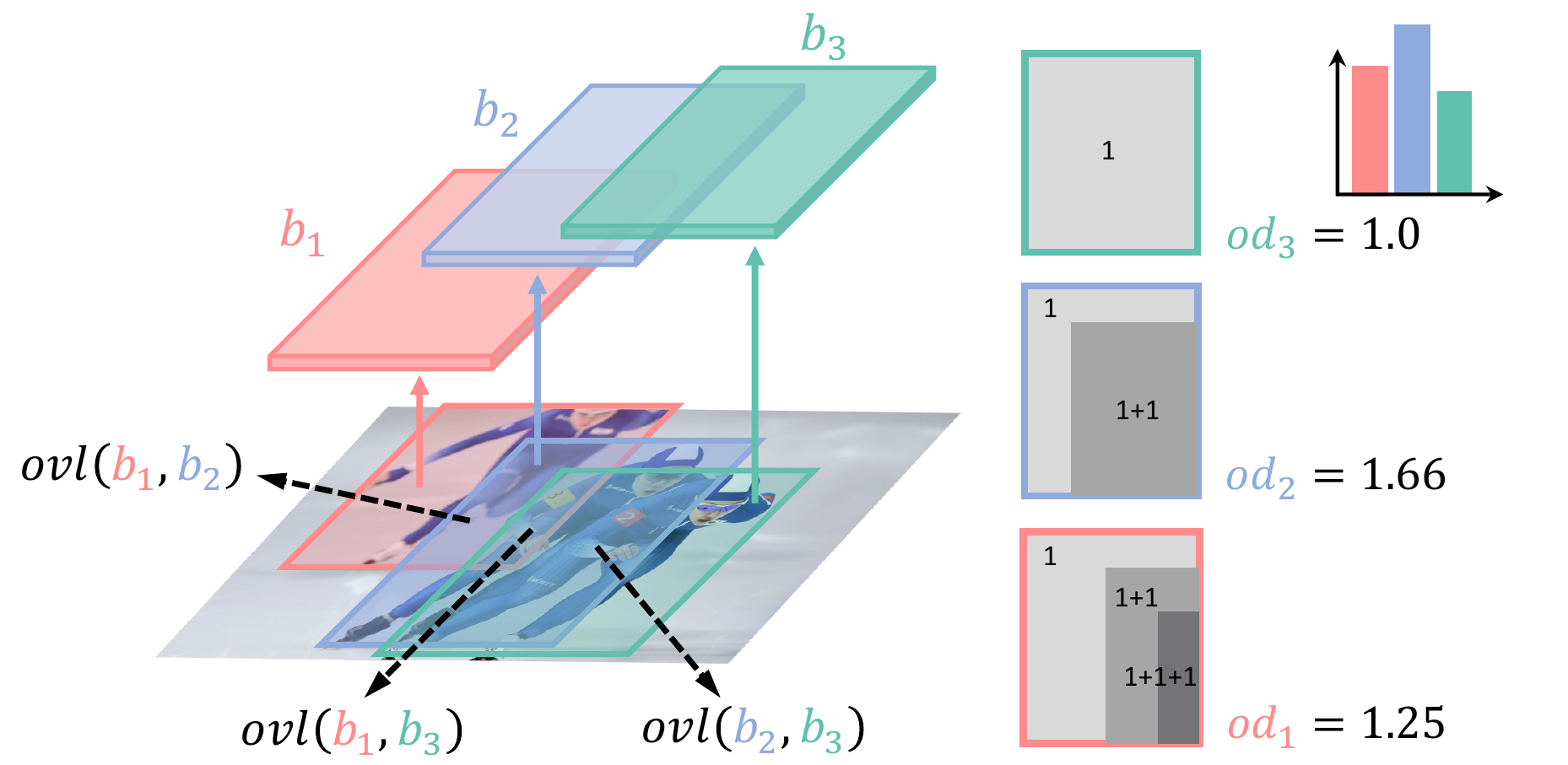}
  \vspace{-1.5\baselineskip}
  \caption{{\bf{Definition of overlay depth (OD).}}
            Calculation process of the OD value as defined in Eq.(\ref{eq:od}).
            Boxes of $b_1$, $b_2$ and $b_3$ are three overlapped objects (skaters),
            in which $b_2$ is overlaid by $b_3$ only while $b_1$ is overlaid by both $b_2$ and $b_3$.
            }
\label{fig:od}
\end{figure}

First, we introduce a variable named ``overlay depth'' (OD) that depicts the extent of how an object is visually overlaid by others.
Fig.~\ref{fig:od} demonstrates the process of calculating OD.
We start by assuming that the overlay depth of an object equals to 1.0 if there are no other objects covering it.
Let $ovl(b_1, b_2)$ be the region of object $b_1$ overlaid by object $b_2$ and $S(\cdot)$ denote the size of a region.
For any object $b_i$ in the image, there exists a set $\mathcal{O}_i$ of objects overlying $b_i$:
\begin{equation}
\label{eq:O_i}
    \mathcal{O}_i = \{b_j\in\mathcal{B} | b_j\neq b_i, S(ovl(b_i, b_j))>0\},
\end{equation}
where $\mathcal{B}$ is the set of all objects in current image.
Then, the OD value of $b_i$ can be clearly defined:
\begin{equation}
\label{eq:od}{}
    od_i = 1.0 + \frac{1}{S(b_i)}\sum_{b_j\in\mathcal{O}_i}S(ovl(b_i, b_j)).
\end{equation}
Therefore, the severer an object is occluded by others (objects of the same category),
the higher OD value it would be assigned (such as objects $b_1$ and $b_2$ in Fig.~\ref{fig:od}).
Starting from this property, application of the overlay depth is based on a plausible observation:
two heavily overlapped objects usually lie in different depth, or more specifically, hold distinct OD values. 
So by taking extra knowledge from the axis of depth,
the OD value can be adopted during de-duplication in a confused 2D plane.
%
%

Now we enable the detector to predict the OD values. 
Generally, a detection model takes a branch to regress the coordinates of the bounding-box.
Following this design, we add an extra predictor to the branch in taking responsibility for the OD regression.
This modification incurs neglectable computing burden and can be easily implemented in both one-stage and two-stage structures (refer to the Appendix for details).
During training, a common L2 loss is adopted.
It should be emphasized that only the OD of pasted objects can be acquired due to the semi-supervised knowledge of the overlay depth.
So we activate the OD regression loss only when the ground-truth is available.
Formally, the whole loss can be written as below:
\begin{equation}
\label{eq:loss}
    L_{det} =
    \begin{cases}
    \alpha\cdot L_{cls\_reg} + \gamma\cdot L_{cl} + \eta\cdot L_{od}& \text{if \emph{od} available}\\
    \alpha\cdot L_{cls\_reg} + \gamma\cdot L_{cl}& \text{elsewise},
    \end{cases}
\end{equation}
where $L_{cls\_reg}$ is the conventional detection loss, $L_{cl}$ is the consensus learning loss and $L_{od}$ is OD regression loss respectively.
We use $\alpha=\gamma=1$ and $\eta=0.1$ in this paper.
%
%
%
\begin{algorithm}[h]
    \caption{Overlay Depth-aware NMS}
    \label{od-nms}
    \begin{algorithmic}
        \STATE \textbf{Input}: $\mathcal{B}=\{b_1,...,b_N\}$: All boxes; $\mathcal{S}=\{s_1,...,s_N\}$: Scores; $th_{iou}$: IoU threshold.\\
        \STATE $\mathcal{D} \gets \varnothing$
        \WHILE{$\mathcal{B} \neq \varnothing$}
            \STATE $m \gets argmax\{\mathcal{S}\}$
            \STATE $\mathcal{M} \gets b_m$; $\mathcal{D} \gets \mathcal{D} \bigcup \mathcal{M}; \mathcal{B} \gets \mathcal{B}-\mathcal{M}$
            \FOR{$b_i$ in $\mathcal{B}$}
                \STATE $th_{od} = \delta \cdot e^{\psi \cdot IoU(\mathcal{M}, b_i)}$
                \IF{$IoU(\mathcal{M}, b_i) \geqslant th_{iou}$ \bf{and} $|od_i - od_m| \leqslant th_{od}$}
                    \STATE $\mathcal{B} \gets \mathcal{B}-b_i; \mathcal{S} \gets \mathcal{S}-s_i$
                \ENDIF
            \ENDFOR
        \ENDWHILE
    \end{algorithmic}
\end{algorithm}

During inference, we invent a novel de-duplication strategy named Overlay Depth-aware NMS (OD-NMS).
In the original NMS pipeline, boxes are recursively compared with each other and one of them would be suppressed in each step if the IoU exceeds a threshold $th_{iou}$.
Following this scheme, objects might be de-duplicated by mistake in a crowded scenario.
In our OD-NMS, for difficult scenario where IoU is higher than $th_{iou}$, we integrate the predicted OD value into a more comprehensive decision.
If the two objects are in different depth, \emph{i.e.}, the absolute difference of the two OD values is higher than a predefined threshold $th_{od}$, we can cancel the suppression in the current step.
Empirically, ambiguous cases often raise in the range of large IoU:
when two boxes are more heavily overlapped, we need stricter OD threshold to judge if they are distinct objects.
So we design a dynamic threshold of OD with respect to the IoU value:
\begin{equation}
\label{eq:loss}
  th_{od} = \delta \cdot e^{\psi \cdot IoU},
\end{equation}
where $\delta$ and $\psi$ are constant coefficients.

Algorithm~\ref{od-nms} summarizes the whole process.
In this way, objects in a crowded scenario can be effectively recalled instead of being inappropriately de-duplicated.
This strategy can be viewed as an evolvement of the original NMS with comparable time complexity.

\section{Experiment}
\textbf{{Datasets.}}
Pedestrian detection is the most typical task burdened by the crowdedness problem,
so our experiments are conducted mainly on two datasets: CrowdHuman~\cite{shao2018crowdhuman} and CityPersons~\cite{zhang2017citypersons}.
Annotations in these datasets consist of a full box and a visible box for each person,
in which we only adopt the full ones to make the data crowded enough. 
Since both the training and validation data hold the same level of crowdedness,
we prepare another ``sparse training set'' by re-labeling full body box of persons in COCO~\cite{Lin2014Microsoft} to further evaluate the potential of our method.
We name this train set as COCO-fullperson (we will release this dataset).
Moreover, we use the category of ``car'' in KITTI~\cite{Geiger2012CVPR} to further estimate the generality of our work in other types of objects.
\\[5pt]
\textbf{Augmentation Details.}
For pasting instance generation, we choose the open source Mask R-CNN~\cite{he2017mask} model adopting ResNet-50~\cite{he2016deep} as backbone.
We run this model on the train set and select 1000 instances with only three rough criteria: high confidence, relatively large size and not been occluded.
A group of fixed hyper parameters are used in our experiments,
where sample numbers $N=3$ and $M=5$, shifting coefficients $\tau=4$, $\epsilon=2$ and OD-NMS coefficient $\delta=0.001$, $\psi=10$.
Copy-paste augmentation strategies are processed online within each training step,
along with the generation of the semi-supervised OD ground-truths according to Eq.(\ref{eq:od}).
We start consensus learning at the 10-th epoch during training.
\\[5pt]
\textbf{Experimental Settings.}
We conduct experiments on both two-stage and one-stage detection frameworks.
For two-stage structure, we use the standard Faster R-CNN~\cite{Ren2015Faster} with FPN~\cite{lin2017feature}.
For one-stage structure, we choose RetinaNet~\cite{Lin2017Focal} as a representative.
All those detectors use ResNet-50 as backbone.
We train the networks on 8 Nvidia V100 GPUs with 2 images on each GPU.
We also apply our method to the state-of-the-art pedestrian detectors CrowdDet~\cite{chu2020detection} and ProgS-RCNN~\cite{zheng2022progressive}.
Other training details will be reported in the following subsections.

\subsection{Results on CrowdHuman}
%
%
\begin{table}[]
\renewcommand\arraystretch{0.8}
\scriptsize
\centering
\begin{tabular}{l|cccc}
\toprule
                    & $MR^{-2}$  & AP@0.5    & AP@0.5:0.95 & JI \\ \hline
\makecell{Aug Method}     & \multicolumn{4}{c}{on \emph{Faster R-CNN}} \\ \hline
Baseline             & 50.42      &84.95      &-            &-        \\
Baseline$^+$         & 42.46      &87.07      &52.70        &79.77    \\
Mosaic               & 43.71      &85.21      &52.66        &78.35    \\
RandAug              & 42.17      &87.48      &53.19        &80.40    \\
SAutoAug             & 42.13      &87.64      &53.35        &80.39    \\
SimCP                & 41.88      &87.36      &53.36        &79.53    \\
CrowdAug (\textbf{Ours}) & \textbf{40.21}      &\textbf{88.61}      &\textbf{54.88}        &\textbf{81.41} \\ \hline
\makecell{Aug Method}     & \multicolumn{4}{c}{on \emph{RetinaNet}} \\ \hline
Baseline             & 63.33     &80.83    &-            &-      \\
Baseline$^+$         & 50.65     &83.80    &49.63        &76.40  \\
Mosaic               & 52.53     &82.95    &48.87        &75.60  \\
RandAug              & 50.25     &83.94    &49.77        &76.58  \\
SAutoAug             & 50.21     &84.02    &49.85        &76.80  \\
SimCP                & 50.01     &84.12    &50.05        &77.02  \\
CrowdAug (\textbf{Ours}) & \textbf{47.35}  &\textbf{85.29} &\textbf{51.84}  &\textbf{77.79}  \\ \hline
\multicolumn{5}{c}{on \emph{SOTA} pedestrian detectors} \\ \hline
CrowdDet            & 41.35      &90.06      &55.02        &82.07   \\
ProgS-RCNN          & 41.45      &92.15      &58.17        &83.13   \\
CrowdDet + AutoPedestrian & 40.58 &-         &-             &-         \\
CrowdDet + \textbf{Ours} & \textbf{38.98}      &91.50    &57.65        &\textbf{83.89}   \\
ProgS-RCNN + \textbf{Ours} & 40.12         & \textbf{92.31}        &\textbf{58.20}             & 83.35        \\ 
\bottomrule
\end{tabular}
  \caption{\small{
    \textbf{Results on CrowdHuman val set.}
    The Baseline$^+$ denotes newly trained strong baselines.
    Results are in percentage (\%).
  }}
\label{tab:crowdhuman}
\end{table}
%
Four metrics are used to evaluate results on CrowdHuman:
the \emph{log-average miss rate on False Positive Per Image} (FPPI) in the range of $[10^{-2}, 10^0]$ (shortened as MR$^{-2}$, lower is better), the \emph{Average Precisions} (AP@0.5 and AP@0.5:0.95, higher is better) and the \emph{Jaccard Index} (JI, higher is better),
among which the MR$^{-2}$ is the main indicator.
To make our experiments convincing enough, we use very strong baselines (the Baseline$^+$s in Table~\ref{tab:crowdhuman}), which are 8\%-12\% superior than those in the CrowdHuman paper~\cite{shao2018crowdhuman}.
During training, the short side of each image is resized to 800 and the long side is limited within 1400.
Models are trained for 60k iterations starting from an initial learning rate of 0.02 (Faster R-CNN) or 0.01 (RetinaNet) and is reduced by 0.1 on 30k and 40k iters respectively.
Table~\ref{tab:crowdhuman} compares results of our method (CrowdAug) with other approaches.
First, the widely used Mosaic augmentation~\cite{DBLP:journals/corr/abs-2004-10934} leads to a decline.
This phenomenon is mainly attributed to the fact that in CrowdHuman, many boxes extend across image boundary.
After the mosaic operation, these near-boundary boxes are truncated at the joints of image patches, losing original characteristics.
We also make trials of two automated strategies:
the Random-Augmentation (RandAug)~\cite{DBLP:conf/nips/CubukZS020} and the Scale-Aware Auto-Augmentation (SAutoAug)~\cite{chen2021scale}.
It needs to be noted that in these works, the search space does not include policies in dealing with crowded scene,
which we hypothesize is the main reason of their marginal effects.
The Simple Copy-Paste~\cite{DBLP:conf/cvpr/GhiasiCSQLCLZ21} (SimCP in Table~\ref{tab:crowdhuman})improves the detector by nearly 0.6\%.
Instead, our CrowdAug can consistently improve the detection results by 2.2\% and 3.3\% for Faster R-CNN and RetinaNet respectively from the strong baselines.
Moreover, the proposed method has exceptional performance on the state-of-the-art (SOTA) pedestrian detectors CrowdDet~\cite{chu2020detection} and ProgS-RCNN~\cite{zheng2022progressive}.
As shown in the last two lines of Table~\ref{tab:crowdhuman}.
On CrowdDet, our method can achieve an improvement of 2.37\% and reach a new SOTA of \textbf{38.98\%} in MR$^{-2}$.
On ProgS-RCNN (only the CCP is applied since the CL and OD-NMS is not needed for end-to-end detector),
our method can bring an enhancement of 1.33\%.
The proposed CrowdAug can also outperform the previously SOTA augmentation strategy AutoPedestrian~\cite{tang2021autopedestrian} by 1.6\% in MR$^{-2}$.
These experiments confirm that the CrowdAug can effectively optimize the crowded detection even on a supremely high base.

We also train the detector on the ``sparse'' dataset COCO-fullperson and report results on the ``crowded'' CrowdHuman val set in Table~\ref{tab:coco-crowdhuman}.
Since training samples are generally not crowded,
the CrowdAug can bring significant improvement (more than 3\% in MR$^{-2}$). 
These results suggest that our method can largely help the detector to handle crowded scenes when there is limited or even no crowded data available for training.
%
%
\begin{table}[]
\renewcommand\arraystretch{0.8}
\scriptsize
\centering
\begin{tabular}{l|cccc}
\hline
& MR$^{-2}$  & AP@0.5    & AP@0.5:0.95 & JI \\ \hline
Faster R-CNN     & 53.51      &85.30      &46.33        &77.21    \\
Faster R-CNN + \textbf{Ours} &\textbf{50.12}      &\textbf{86.40}      &\textbf{48.52}   &\textbf{78.50} \\\hline
RetinaNet     & 59.45     &80.86    &41.71        &74.22  \\
RetinaNet + \textbf{Ours} &\textbf{56.80}     &\textbf{81.42}    &\textbf{43.41}        &\textbf{75.30}  \\ \hline
\end{tabular}
  \caption{\small{
    Results of model trained on COCO-fullperson and evaluated on CrowdHuman val set. We list results on Faster R-CNN and RetinaNet respectively.
  }}
\label{tab:coco-crowdhuman}
\end{table}
%

\subsection{Ablation Study}
\begin{figure}[t]
  \centering
  \includegraphics[width=1.\linewidth]{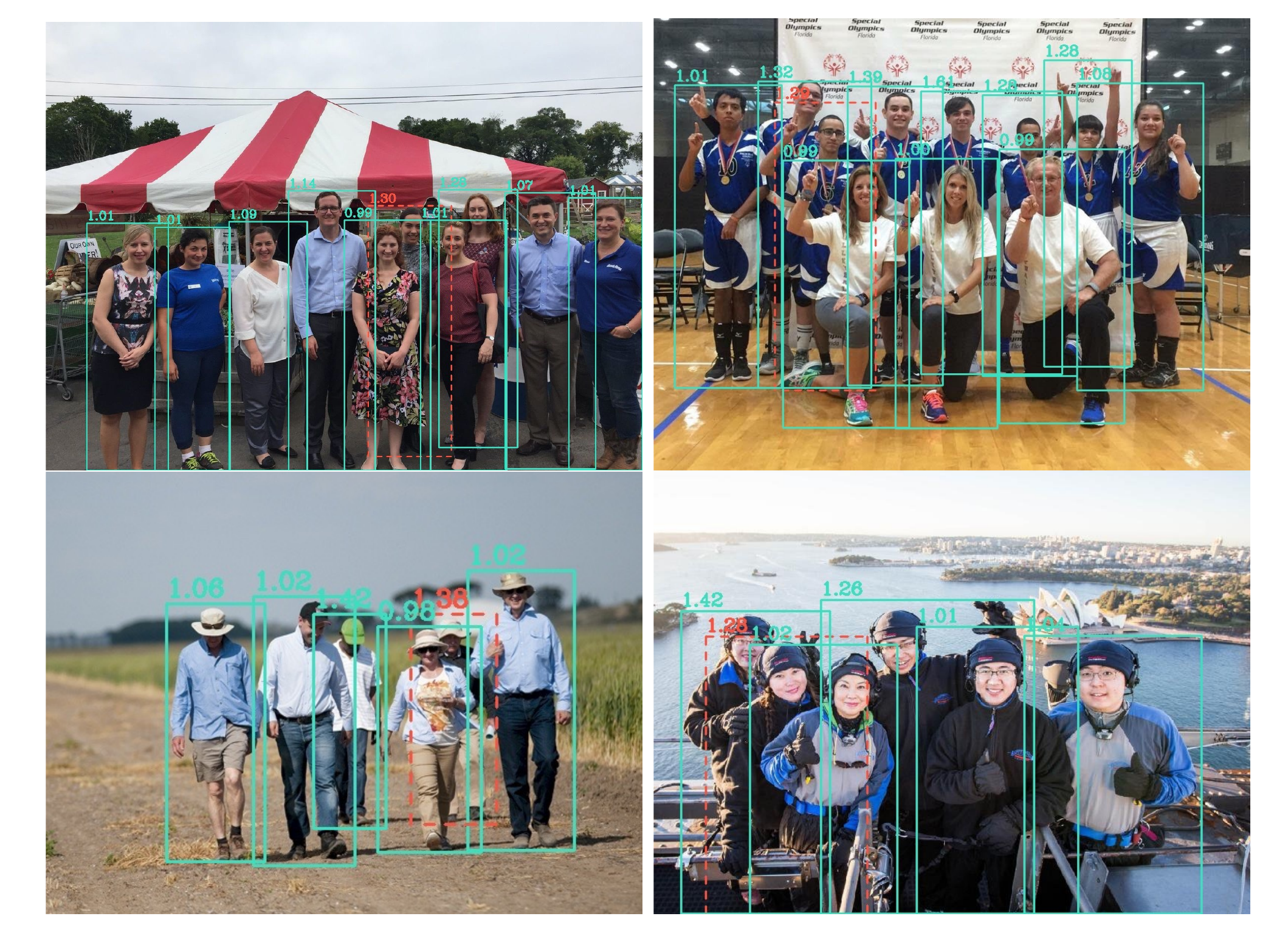}
  \caption{{\bf{Visualization of the OD prediction.}}
            The value of predicted overlay depth (OD) is marked at the top left corner of each box.
            The \textit{red} boxes denote the persons who are wrongly deleted by the original NMS while recalled by our OD-NMS.
            }
\label{fig:od-visualization}
\end{figure}

\textbf{Crowdedness-oriented Design.}
The third line of Table~\ref{tab:ablation} shows the contribution of our augmentation strategy (CCP).
Take the Faster R-CNN as an example.
The CCP can improve the detection result by nearly 1.3\%.
For comparison, we try the random copy-paste (RCP) mentioned in Sec.\ref{sec:resist-icd}.
In this strategy, average number and size distribution of pasting objects are kept the same with those in our CCP
while the positions to paste are randomly allocated rather than specially making crowded scenes.
The 2nd line of Table~\ref{tab:ablation} shows that the RCP improves the baseline by 0.45\%, which is inferior to our CCP.
These results demonstrate that operations in boosting the crowdedness are necessary and effective.
As discussed in Sec.\ref{sec:resist-icd},
we think the improvement comes mainly from resisting the ICD issue.
%
%
\begin{table}[]
\renewcommand\arraystretch{0.8}
\tabcolsep=0.27cm
\scriptsize
\begin{tabular}{ccc|cc|cc}
\hline
&         &        & MR$^{-2}$& AP@0.5                & MR$^{-2}$                  & AP@0.5                  \\ \hline
CCP       & CL     & OD       & \multicolumn{2}{c|}{on \emph{Faster R-CNN}} & \multicolumn{2}{c}{on \emph{RetinaNet}} \\ \hline
          &        &          & 42.46              & 87.07                 & 50.65                & 83.80                   \\
(RCP)     &        &          & 42.01              & 87.10                 & 49.75                & 84.02                   \\
$\surd$   &        &          & 41.11              & 87.75                 & 48.81                & 84.73                   \\
$\surd$   &$\surd$ &          & 40.80              & 88.02                 & 47.93                & 84.85                   \\
$\surd$   &$\surd$ &$\surd$   & 40.21              & 88.61                 & 47.35                & 85.29                   \\ \hline
\end{tabular}
  \caption{\small{
    \textbf{Ablation results on CrowdHuman val set.}
    Experiments are conducted on Faster R-CNN and RetinaNet respectively.
  }}
\label{tab:ablation}
\end{table}
\\[5pt]
\noindent\textbf{Consensus Learning.}
As shown in the 4-th line of Table~\ref{tab:ablation},
the proposed consensus learning (CL) strategy can further enhance the the Faster R-CNN by 0.3\% from CCP baseline.
This improvement becomes much larger (0.88\%) when applying the CL to RetinaNet.
Additionally, with qualitative analysis in Sec.\ref{sec:resist-icd},
we can make a conclusion that this module makes a step further in alleviating the ICD problem.
\\[5pt]
\noindent\textbf{Overlay Depth.}
Comparing the last two lines of Table~\ref{tab:ablation} can find out contribution of the overlay depth (OD).
As a breakthrough of the 2D constraint, this weak depth knowledge brings a stable enhancement.
We make visualizations of the OD prediction in Fig.~\ref{fig:od-visualization}.
It can be seen that although the training process is semi-supervised, overlay depths learned by the detector are quite discriminative and can recall missing pedestrians (red dotted boxes in Fig.~\ref{fig:od-visualization}) of the baseline model.
In the structure design, the simplicity of our OD predictor guarantees the ease of use during application.
%
%
\\[5pt]
\textbf{Robustness to Pasting Objects.} Our method is robust to the quantity and quality of pasting objects.
First, we experiment the CrowdAug on a variety of pasting objects numbers.
Then, we manually select 1000 high-quality object patches and train a comparing model (the 4th line in Table~\ref{tab:p}).
Finally, we replace the mask of the object patches above with the segmentation annotations in COCO~\cite{Lin2014Microsoft} to get more precise masks and make another experiment (the 5th line in Table~\ref{tab:p}).
Results show that variations of either quantity or quality of pasting objects will not essentially effect the final performance,
which suggest that our method does not have strict requirement of the training data and with huge potential of application.
%
%
\begin{table}[]
\renewcommand\arraystretch{0.8}
\scriptsize
\centering
\begin{tabular}{l|cccc}
\hline
\makecell{Pasting Object Numbers}         & MR$^{-2}$  & AP@0.5    & AP@0.5:0.95 & JI  \\ \hline
1000 (default)          & 40.21      &88.61      &54.88        &81.41  \\
3000                    & 40.25      &88.53      &54.85        &81.39  \\
500                     & 40.23      &88.57      &54.88        &81.40  \\
1000 sel                & 40.20      &88.60      &54.90        &81.32  \\
1000 sel$+$mask gt      & 40.21      &88.62      &54.86        &81.42  \\ \hline
\end{tabular}
  \caption{\small{
    \textbf{Robustness to Pasting Objects.}
    The ``sel'' denotes manually selected high-quality objects and the
    ``mask gt'' means using segmentation annotations instead of those predicted by the Mask R-CNN model.
  }}
\label{tab:p}
\end{table}
%

\subsection{Results on CityPersons}
On CityPersons, images are trained and evaluated with input scale of $\times$1.3.
During training, we use an initial learning rate of 0.02 (Faster R-CNN) or 0.01 (RetinaNet) for the first 5k iterations and reduce it by 0.1 continuously on the next two groups of 2k iterations.
Table~\ref{tab:citypersons} compares our CrowdAug with other methods.
The results show that the CrowdAug can stably optimize the detector and once the crowdedness becomes heavier, the improvement becomes larger.
Meanwhile, the proposed method can also outperform other counterparts.
%
%
\begin{table}[]
\renewcommand\arraystretch{0.8}
\tabcolsep=0.13cm
\scriptsize
\centering
\begin{tabular}{l|ccccc}
\hline
\multirow{3}{*}{\centering{Method}} & \multicolumn{4}{c|}{MR$^{-2}$}                                & \multirow{2}{*}{AP@0.5}  \\ \cline{2-5} 
                        & Reasonable & Partial & Bare & \multicolumn{1}{l|}{Heavy} &                         \\ \hline
Faster R-CNN                &11.20      &11.55      &6.62       &52.05       &82.95      \\
Faster R-CNN + Mosaic                  &11.05      &11.42      &6.77       &51.62       &83.01      \\
Faster R-CNN + RandAug                 &10.84      &11.20      &6.31       &51.27       &82.97      \\
Faster R-CNN + APGAN          &11.9      &11.9      &6.8       &49.6       &-      \\
Faster R-CNN + AutoPedestrian          &10.3      &-      &-       &49.4       &-      \\
Faster R-CNN + \textbf{Ours}                &\textbf{10.02}      &\textbf{10.48}      &\textbf{5.79}       &\textbf{48.50}       &\textbf{83.78} \\ \hline
RetinaNet                &13.60        &14.32        &7.22       &55.61                   &79.31                         \\
RetinaNet + Mosaic                  &13.20        &14.58        &7.50       &54.90                   &79.31                         \\
RetinaNet + RandAug                 &13.23        &13.96        &7.02       &54.61                   &79.77                         \\
RetinaNet + \textbf{Ours}           &\textbf{12.38}        &\textbf{13.07}        &\textbf{6.49}       &\textbf{52.96}                   &\textbf{80.86} \\ \hline
\end{tabular}
  \caption{\small{
    \textbf{Results on CityPersons val set.}
    We list the MR$^{-2}$ on four crowdedness levels: \emph{reasonable}, \emph{partial}, \emph{bare} and \emph{heavy}. The metric of AP@0.5 is also reported.
  }}
\label{tab:citypersons}
\end{table}

\subsection{Results on KITTI}
To estimate the generalization of our method to other crowded objects, we make experiments on the category of ``cars'' in KITTI~\cite{Geiger2012CVPR}.
Table~\ref{tab:kitti} shows the results.
After applying the CrowdAug, Average Precision of cars get improvement if 1.05\%, 1.20\% and 2.25\% for the objects of easy, moderate and hard respectively for the Faster R-CNN structure,
which demonstrate the similar trend of its performance on pedestrian detection.
%
%
\begin{table}[]
\tabcolsep=0.2cm
\scriptsize
\centering
\begin{tabular}{l|ccc|ccc}
\hline
\multirow{2}{*}{}  & \multicolumn{1}{l|}{Easy} & \multicolumn{1}{l|}{Moderate} & Hard & \multicolumn{1}{l|}{Easy} & \multicolumn{1}{l|}{Moderate} & Hard \\ \cline{2-7}
                   & \multicolumn{3}{c|}{on \emph{Faster R-CNN}}                                & \multicolumn{3}{c}{on \emph{RetinaNet}}                                    \\ \hline
Baseline           &97.24          &89.77           &79.44           &93.72           &87.33           &76.76      \\
CrowdAug           &\textbf{98.30}          &\textbf{91.07}           &\textbf{81.69}           &\textbf{94.81}           &\textbf{88.59}           &\textbf{78.63}      \\ \hline
\end{tabular}
  \caption{\small{
    \textbf{Results on KITTI val set.}
    We use the category of ``cars'' in KITTI~\cite{Geiger2012CVPR} dataset.
    AP@0.7 (\%) of \emph{easy}, \emph{moderate} and \emph{hard} objects are listed respectively.
  }}
\label{tab:kitti}
\end{table}

\section{Conclusion}
In this paper, we point out two main effects of crowdedness issue in the visual object detection task
and propose a solution from the perspective of data augmentation.
First, we invent a novel copy-paste strategy to improve crowdedness and design a consensus learning method.
Then, we reasonably use the weak information of depth produced by the pasting process.
Both contributions can help alleviating the ambiguities of crowded 2D object detection.
We think this is a new pathway of solving the crowdedness issue with the advantages of significant effect and resource conservation.

\bibliography{aaai23}

\begin{thebibliography}{46}
\providecommand{\natexlab}[1]{#1}

\bibitem[{Bochkovskiy, Wang, and
  Liao(2020)}]{DBLP:journals/corr/abs-2004-10934}
Bochkovskiy, A.; Wang, C.; and Liao, H.~M. 2020.
\newblock YOLOv4: Optimal Speed and Accuracy of Object Detection.
\newblock \emph{CoRR}, abs/2004.10934.

\bibitem[{Chen et~al.(2021)Chen, Li, Kong, Qi, Chu, Li, and
  Jia}]{chen2021scale}
Chen, Y.; Li, Y.; Kong, T.; Qi, L.; Chu, R.; Li, L.; and Jia, J. 2021.
\newblock Scale-aware automatic augmentation for object detection.
\newblock In \emph{Proceedings of the IEEE/CVF Conference on Computer Vision
  and Pattern Recognition}, 9563--9572.

\bibitem[{Chen et~al.(2020)Chen, Zhang, Li, Li, Zhang, Meng, Xiang, Sun, and
  Jia}]{DBLP:journals/corr/abs-2004-12432}
Chen, Y.; Zhang, P.; Li, Z.; Li, Y.; Zhang, X.; Meng, G.; Xiang, S.; Sun, J.;
  and Jia, J. 2020.
\newblock Stitcher: Feedback-driven Data Provider for Object Detection.
\newblock \emph{CoRR}, abs/2004.12432.

\bibitem[{Chu et~al.(2020)Chu, Zheng, Zhang, and Sun}]{chu2020detection}
Chu, X.; Zheng, A.; Zhang, X.; and Sun, J. 2020.
\newblock Detection in crowded scenes: One proposal, multiple predictions.
\newblock In \emph{Proceedings of the IEEE/CVF Conference on Computer Vision
  and Pattern Recognition}, 12214--12223.

\bibitem[{Cubuk et~al.(2018)Cubuk, Zoph, Mane, Vasudevan, and
  Le}]{2018AutoAugment}
Cubuk, E.; Zoph, B.; Mane, D.; Vasudevan, V.; and Le, Q.~V. 2018.
\newblock AutoAugment: Learning Augmentation Policies from Data.

\bibitem[{Cubuk et~al.(2020)Cubuk, Zoph, Shlens, and
  Le}]{DBLP:conf/nips/CubukZS020}
Cubuk, E.~D.; Zoph, B.; Shlens, J.; and Le, Q. 2020.
\newblock RandAugment: Practical Automated Data Augmentation with a Reduced
  Search Space.
\newblock In \emph{Advances in Neural Information Processing Systems 33: Annual
  Conference on Neural Information Processing Systems 2020, NeurIPS 2020,
  December 6-12, 2020, virtual}.

\bibitem[{Devries and Taylor(2017)}]{DBLP:journals/corr/abs-1708-04552}
Devries, T.; and Taylor, G.~W. 2017.
\newblock Improved Regularization of Convolutional Neural Networks with Cutout.
\newblock \emph{CoRR}, abs/1708.04552.

\bibitem[{Dvornik, Mairal, and Schmid(2018)}]{dvornik2018modeling}
Dvornik, N.; Mairal, J.; and Schmid, C. 2018.
\newblock Modeling visual context is key to augmenting object detection
  datasets.
\newblock In \emph{Proceedings of the European Conference on Computer Vision
  (ECCV)}, 364--380.

\bibitem[{Dwibedi, Misra, and Hebert(2017)}]{DBLP:conf/iccv/DwibediMH17}
Dwibedi, D.; Misra, I.; and Hebert, M. 2017.
\newblock Cut, Paste and Learn: Surprisingly Easy Synthesis for Instance
  Detection.
\newblock In \emph{{IEEE} International Conference on Computer Vision, {ICCV}
  2017, Venice, Italy, October 22-29, 2017}, 1310--1319. {IEEE} Computer
  Society.

\bibitem[{Fang et~al.(2019)Fang, Sun, Wang, Gou, Li, and
  Lu}]{DBLP:conf/iccv/FangSWGLL19}
Fang, H.; Sun, J.; Wang, R.; Gou, M.; Li, Y.; and Lu, C. 2019.
\newblock InstaBoost: Boosting Instance Segmentation via Probability Map Guided
  Copy-Pasting.
\newblock In \emph{2019 {IEEE/CVF} International Conference on Computer Vision,
  {ICCV} 2019, Seoul, Korea (South), October 27 - November 2, 2019}, 682--691.
  {IEEE}.

\bibitem[{G{\"a}hlert et~al.(2020)G{\"a}hlert, Hanselmann, Franke, and
  Denzler}]{gahlert2020visibility}
G{\"a}hlert, N.; Hanselmann, N.; Franke, U.; and Denzler, J. 2020.
\newblock Visibility guided nms: Efficient boosting of amodal object detection
  in crowded traffic scenes.
\newblock \emph{arXiv preprint arXiv:2006.08547}.

\bibitem[{Geiger, Lenz, and Urtasun(2012)}]{Geiger2012CVPR}
Geiger, A.; Lenz, P.; and Urtasun, R. 2012.
\newblock Are we ready for Autonomous Driving? The KITTI Vision Benchmark
  Suite.
\newblock In \emph{Conference on Computer Vision and Pattern Recognition
  (CVPR)}.

\bibitem[{Ghiasi et~al.(2021)Ghiasi, Cui, Srinivas, Qian, Lin, Cubuk, Le, and
  Zoph}]{DBLP:conf/cvpr/GhiasiCSQLCLZ21}
Ghiasi, G.; Cui, Y.; Srinivas, A.; Qian, R.; Lin, T.; Cubuk, E.~D.; Le, Q.~V.;
  and Zoph, B. 2021.
\newblock Simple Copy-Paste Is a Strong Data Augmentation Method for Instance
  Segmentation.
\newblock In \emph{{IEEE} Conference on Computer Vision and Pattern
  Recognition, {CVPR} 2021, virtual, June 19-25, 2021}, 2918--2928. Computer
  Vision Foundation / {IEEE}.

\bibitem[{He et~al.(2017)He, Gkioxari, Doll{\'a}r, and Girshick}]{he2017mask}
He, K.; Gkioxari, G.; Doll{\'a}r, P.; and Girshick, R. 2017.
\newblock Mask r-cnn.
\newblock In \emph{Proceedings of the IEEE international conference on computer
  vision}, 2961--2969.

\bibitem[{He et~al.(2016)He, Zhang, Ren, and Sun}]{he2016deep}
He, K.; Zhang, X.; Ren, S.; and Sun, J. 2016.
\newblock Deep residual learning for image recognition.
\newblock In \emph{Proceedings of the IEEE conference on computer vision and
  pattern recognition}, 770--778.

\bibitem[{Huang et~al.(2020)Huang, Yue, Deng, and Zhou}]{huang2020visible}
Huang, Z.; Yue, K.; Deng, J.; and Zhou, F. 2020.
\newblock Visible Feature Guidance for Crowd Pedestrian Detection.
\newblock \emph{arXiv preprint arXiv:2008.09993}.

\bibitem[{Krizhevsky, Sutskever, and
  Hinton(2012)}]{DBLP:conf/nips/KrizhevskySH12}
Krizhevsky, A.; Sutskever, I.; and Hinton, G.~E. 2012.
\newblock ImageNet Classification with Deep Convolutional Neural Networks.
\newblock In \emph{Advances in Neural Information Processing Systems 25: 26th
  Annual Conference on Neural Information Processing Systems 2012. Proceedings
  of a meeting held December 3-6, 2012, Lake Tahoe, Nevada, United States},
  1106--1114.

\bibitem[{LeCun et~al.(1998)LeCun, Bottou, Bengio, and
  Haffner}]{DBLP:journals/pieee/LeCunBBH98}
LeCun, Y.; Bottou, L.; Bengio, Y.; and Haffner, P. 1998.
\newblock Gradient-based learning applied to document recognition.
\newblock \emph{Proc. {IEEE}}, 86(11): 2278--2324.

\bibitem[{Li et~al.(2021)Li, Sohn, Yoon, and Pfister}]{DBLP:conf/cvpr/LiSYP21}
Li, C.; Sohn, K.; Yoon, J.; and Pfister, T. 2021.
\newblock CutPaste: Self-Supervised Learning for Anomaly Detection and
  Localization.
\newblock In \emph{{IEEE} Conference on Computer Vision and Pattern
  Recognition, {CVPR} 2021, virtual, June 19-25, 2021}, 9664--9674. Computer
  Vision Foundation / {IEEE}.

\bibitem[{Lim et~al.(2019)Lim, Kim, Kim, Kim, and
  Kim}]{DBLP:conf/nips/LimKKKK19}
Lim, S.; Kim, I.; Kim, T.; Kim, C.; and Kim, S. 2019.
\newblock Fast AutoAugment.
\newblock In \emph{Advances in Neural Information Processing Systems 32: Annual
  Conference on Neural Information Processing Systems 2019, NeurIPS 2019,
  December 8-14, 2019, Vancouver, BC, Canada}, 6662--6672.

\bibitem[{Lin et~al.(2017{\natexlab{a}})Lin, Doll{\'a}r, Girshick, He,
  Hariharan, and Belongie}]{lin2017feature}
Lin, T.-Y.; Doll{\'a}r, P.; Girshick, R.; He, K.; Hariharan, B.; and Belongie,
  S. 2017{\natexlab{a}}.
\newblock Feature pyramid networks for object detection.
\newblock In \emph{Proceedings of the IEEE conference on computer vision and
  pattern recognition}, 2117--2125.

\bibitem[{Lin et~al.(2017{\natexlab{b}})Lin, Goyal, Girshick, He, and
  Dollar}]{Lin2017Focal}
Lin, T.~Y.; Goyal, P.; Girshick, R.; He, K.; and Dollar, P. 2017{\natexlab{b}}.
\newblock Focal loss for dense object detection.
\newblock \emph{IEEE Transactions on Pattern Analysis \& Machine Intelligence},
  PP(99): 2999--3007.

\bibitem[{Lin et~al.(2014)Lin, Maire, Belongie, Hays, Perona, Ramanan, Dollár,
  and Zitnick}]{Lin2014Microsoft}
Lin, T.~Y.; Maire, M.; Belongie, S.; Hays, J.; Perona, P.; Ramanan, D.;
  Dollár, P.; and Zitnick, C.~L. 2014.
\newblock Microsoft COCO: Common Objects in Context.
\newblock 8693: 740--755.

\bibitem[{Liu et~al.(2020{\natexlab{a}})Liu, Ouyang, Wang, Fieguth, Chen, Liu,
  and Pietik{\"{a}}inen}]{LiuOWFCLP20}
Liu, L.; Ouyang, W.; Wang, X.; Fieguth, P.~W.; Chen, J.; Liu, X.; and
  Pietik{\"{a}}inen, M. 2020{\natexlab{a}}.
\newblock Deep Learning for Generic Object Detection: {A} Survey.
\newblock \emph{Int. J. Comput. Vis.}, 128(2): 261--318.

\bibitem[{Liu et~al.(2020{\natexlab{b}})Liu, Guo, Hu, Zhao, Zhao, Wang, Zhu,
  Wang, and Tang}]{liu2020novel}
Liu, S.; Guo, H.; Hu, J.-G.; Zhao, X.; Zhao, C.; Wang, T.; Zhu, Y.; Wang, J.;
  and Tang, M. 2020{\natexlab{b}}.
\newblock A novel data augmentation scheme for pedestrian detection with
  attribute preserving GAN.
\newblock \emph{Neurocomputing}, 401: 123--132.

\bibitem[{Liu, Huang, and Wang(2019)}]{liu2019adaptive}
Liu, S.; Huang, D.; and Wang, Y. 2019.
\newblock Adaptive nms: Refining pedestrian detection in a crowd.
\newblock In \emph{Proceedings of the IEEE/CVF Conference on Computer Vision
  and Pattern Recognition}, 6459--6468.

\bibitem[{Liu et~al.(2016)Liu, Anguelov, Erhan, Szegedy, Reed, Fu, and
  Berg}]{liu2016ssd}
Liu, W.; Anguelov, D.; Erhan, D.; Szegedy, C.; Reed, S.; Fu, C.-Y.; and Berg,
  A.~C. 2016.
\newblock Ssd: Single shot multibox detector.
\newblock In \emph{European conference on computer vision}, 21--37. Springer.

\bibitem[{Redmon et~al.(2016)Redmon, Divvala, Girshick, and
  Farhadi}]{redmon2016you}
Redmon, J.; Divvala, S.; Girshick, R.; and Farhadi, A. 2016.
\newblock You only look once: Unified, real-time object detection.
\newblock In \emph{Proceedings of the IEEE conference on computer vision and
  pattern recognition}, 779--788.

\bibitem[{Remez, Huang, and Brown(2018)}]{DBLP:conf/eccv/RemezHB18}
Remez, T.; Huang, J.; and Brown, M. 2018.
\newblock Learning to Segment via Cut-and-Paste.
\newblock In Ferrari, V.; Hebert, M.; Sminchisescu, C.; and Weiss, Y., eds.,
  \emph{Computer Vision - {ECCV} 2018 - 15th European Conference, Munich,
  Germany, September 8-14, 2018, Proceedings, Part {VII}}, volume 11211 of
  \emph{Lecture Notes in Computer Science}, 39--54. Springer.

\bibitem[{Ren et~al.(2015)Ren, He, Girshick, and Sun}]{Ren2015Faster}
Ren, S.; He, K.; Girshick, R.; and Sun, J. 2015.
\newblock Faster R-CNN: towards real-time object detection with region proposal
  networks.
\newblock In \emph{International Conference on Neural Information Processing
  Systems}, 91--99.

\bibitem[{Shao et~al.(2018)Shao, Zhao, Li, Xiao, Yu, Zhang, and
  Sun}]{shao2018crowdhuman}
Shao, S.; Zhao, Z.; Li, B.; Xiao, T.; Yu, G.; Zhang, X.; and Sun, J. 2018.
\newblock CrowdHuman: A Benchmark for Detecting Human in a Crowd.
\newblock \emph{arXiv preprint arXiv:1805.00123}.

\bibitem[{Shorten and Khoshgoftaar(2019)}]{ShortenK19}
Shorten, C.; and Khoshgoftaar, T.~M. 2019.
\newblock A survey on Image Data Augmentation for Deep Learning.
\newblock \emph{J. Big Data}, 6: 60.

\bibitem[{Simonyan and Zisserman(2015)}]{DBLP:journals/corr/SimonyanZ14a}
Simonyan, K.; and Zisserman, A. 2015.
\newblock Very Deep Convolutional Networks for Large-Scale Image Recognition.
\newblock In \emph{3rd International Conference on Learning Representations,
  {ICLR} 2015, San Diego, CA, USA, May 7-9, 2015, Conference Track
  Proceedings}.

\bibitem[{Sun et~al.(2021)Sun, Zhang, Jiang, Kong, Xu, Zhan, Tomizuka, Li,
  Yuan, Wang et~al.}]{sun2021sparse}
Sun, P.; Zhang, R.; Jiang, Y.; Kong, T.; Xu, C.; Zhan, W.; Tomizuka, M.; Li,
  L.; Yuan, Z.; Wang, C.; et~al. 2021.
\newblock Sparse r-cnn: End-to-end object detection with learnable proposals.
\newblock In \emph{Proceedings of the IEEE/CVF conference on computer vision
  and pattern recognition}, 14454--14463.

\bibitem[{Szegedy et~al.(2014)Szegedy, Liu, Jia, Sermanet, and
  Rabinovich}]{2014Going}
Szegedy, C.; Liu, W.; Jia, Y.; Sermanet, P.; and Rabinovich, A. 2014.
\newblock Going Deeper with Convolutions.
\newblock \emph{IEEE Computer Society}.

\bibitem[{Tan and Le(2019)}]{tan2019efficientnet}
Tan, M.; and Le, Q. 2019.
\newblock Efficientnet: Rethinking model scaling for convolutional neural
  networks.
\newblock In \emph{International conference on machine learning}, 6105--6114.
  PMLR.

\bibitem[{Tang et~al.(2021)Tang, Li, Liu, Chen, Wang, and
  Ouyang}]{tang2021autopedestrian}
Tang, Y.; Li, B.; Liu, M.; Chen, B.; Wang, Y.; and Ouyang, W. 2021.
\newblock Autopedestrian: an automatic data augmentation and loss function
  search scheme for pedestrian detection.
\newblock \emph{IEEE transactions on image processing}, 30: 8483--8496.

\bibitem[{Wang et~al.(2018)Wang, Xiao, Jiang, Shao, Sun, and
  Shen}]{wang2018repulsion}
Wang, X.; Xiao, T.; Jiang, Y.; Shao, S.; Sun, J.; and Shen, C. 2018.
\newblock Repulsion loss: Detecting pedestrians in a crowd.
\newblock In \emph{Proceedings of the IEEE Conference on Computer Vision and
  Pattern Recognition}, 7774--7783.

\bibitem[{Xie et~al.(2020)Xie, Cholakkal, Anwer, Khan, Pang, Shao, and
  Shah}]{xie2020count}
Xie, J.; Cholakkal, H.; Anwer, R.~M.; Khan, F.~S.; Pang, Y.; Shao, L.; and
  Shah, M. 2020.
\newblock Count-and similarity-aware r-cnn for pedestrian detection.
\newblock In \emph{European Conference on Computer Vision}, 88--104. Springer.

\bibitem[{Yun et~al.()Yun, Han, Chun, Oh, Yoo, and Choe}]{0CutMix}
Yun, S.; Han, D.; Chun, S.; Oh, S.~J.; Yoo, Y.; and Choe, J. ????
\newblock CutMix: Regularization Strategy to Train Strong Classifiers With
  Localizable Features.
\newblock In \emph{International Conference on Computer Vision}.

\bibitem[{Zhang et~al.(2017)Zhang, Cisse, Dauphin, and Lopez-Paz}]{2017mixup}
Zhang, H.; Cisse, M.; Dauphin, Y.~N.; and Lopez-Paz, D. 2017.
\newblock mixup: Beyond Empirical Risk Minimization.

\bibitem[{Zhang, Benenson, and Schiele(2017)}]{zhang2017citypersons}
Zhang, S.; Benenson, R.; and Schiele, B. 2017.
\newblock Citypersons: A diverse dataset for pedestrian detection.
\newblock In \emph{Proceedings of the IEEE Conference on Computer Vision and
  Pattern Recognition}, 3213--3221.

\bibitem[{Zhang et~al.(2018)Zhang, Wen, Bian, Lei, and Li}]{zhang2018occlusion}
Zhang, S.; Wen, L.; Bian, X.; Lei, Z.; and Li, S.~Z. 2018.
\newblock Occlusion-aware R-CNN: detecting pedestrians in a crowd.
\newblock In \emph{Proceedings of the European Conference on Computer Vision
  (ECCV)}, 637--653.

\bibitem[{Zhang et~al.(2021)Zhang, He, Li, Li, See, and
  Lin}]{zhang2021variational}
Zhang, Y.; He, H.; Li, J.; Li, Y.; See, J.; and Lin, W. 2021.
\newblock Variational pedestrian detection.
\newblock In \emph{Proceedings of the IEEE/CVF Conference on Computer Vision
  and Pattern Recognition}, 11622--11631.

\bibitem[{Zheng et~al.(2022)Zheng, Zhang, Zhang, Qi, and
  Sun}]{zheng2022progressive}
Zheng, A.; Zhang, Y.; Zhang, X.; Qi, X.; and Sun, J. 2022.
\newblock Progressive End-to-End Object Detection in Crowded Scenes.
\newblock In \emph{Proceedings of the IEEE/CVF Conference on Computer Vision
  and Pattern Recognition}, 857--866.

\bibitem[{Zoph et~al.(2020)Zoph, Cubuk, Ghiasi, Lin, Shlens, and
  Le}]{DBLP:conf/eccv/ZophCGLSL20}
Zoph, B.; Cubuk, E.~D.; Ghiasi, G.; Lin, T.; Shlens, J.; and Le, Q.~V. 2020.
\newblock Learning Data Augmentation Strategies for Object Detection.
\newblock In \emph{Computer Vision - {ECCV} 2020 - 16th European Conference,
  Glasgow, UK, August 23-28, 2020, Proceedings, Part {XXVII}}, 566--583.
  Springer.

\end{thebibliography}
\clearpage

\appendix
\subsection{Structure of the OD Predictor}
\begin{figure}[t]
\renewcommand{\figurename}{Figure A}
  \centering
  \includegraphics[width=1.\linewidth]{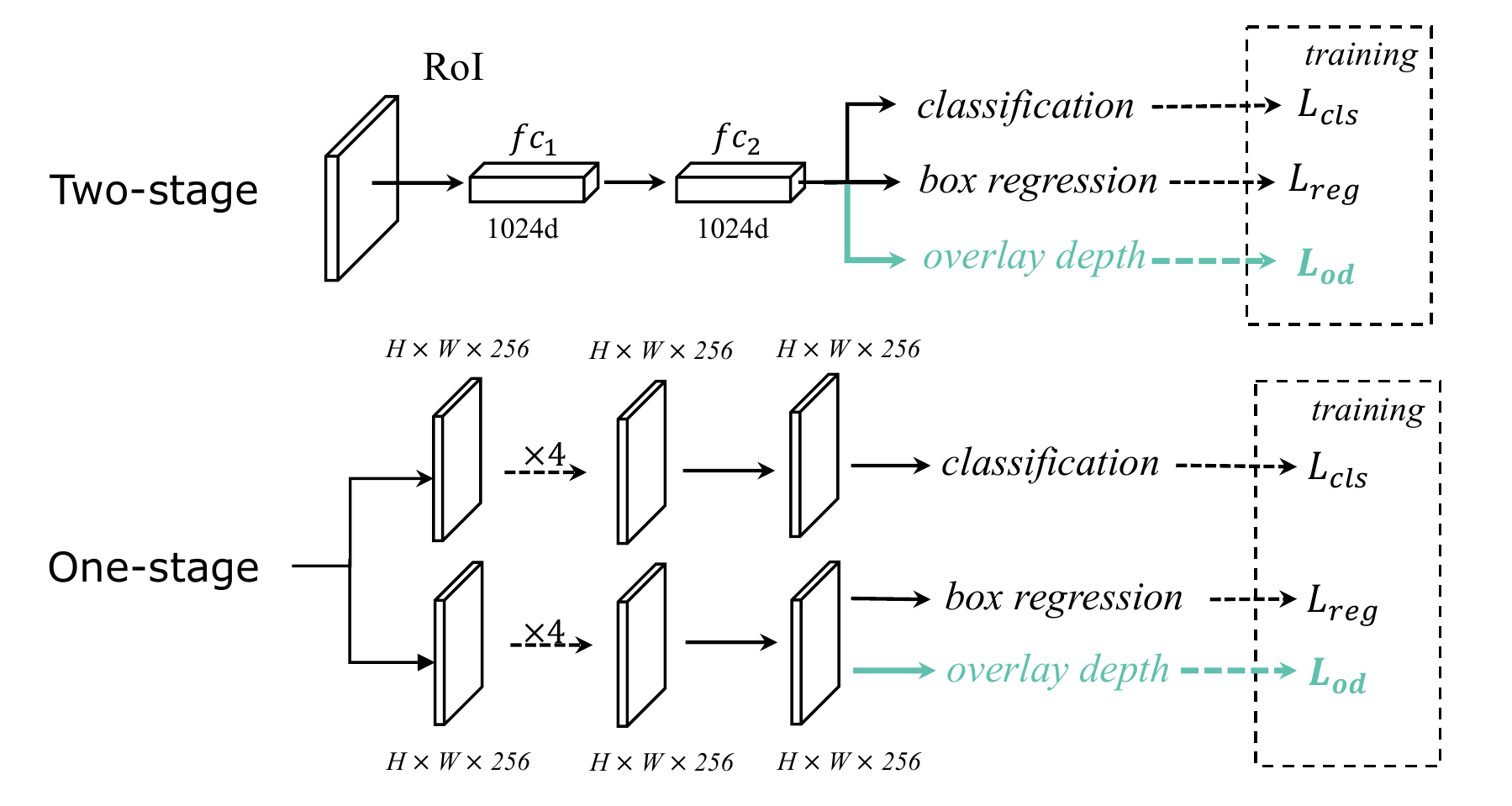}
  \caption{{\bf{Structure of the OD predictor.}}
            The OD predictor (plotted with \textit{green}) on typical two-stage (top) and one-stage (bottom) detection frameworks.
            }
\label{fig:od-detail}
\end{figure}

Fig.A\ref{fig:od-detail} illustrates the details of our overlay depth (OD) predictor.
On two-stage framework, this module is in parallel with the predictor for classification and bounding-box regression,
sharing the same head following RoI-Pooling.
On one-stage framework, the OD predictor is derived from the regression branch.
It is obvious that the newly added module is ultra light-weighted.

\subsection{Implementation Details on CrowdDet}
The CrowdDet (Chu et al. 2020) adopts a multi-instance prediction (MIP) mechanism
to solve the cases in which multiple objects fall into one proposal.
During training, apart from the classification and bounding-box regression,
we add cost functions of consensus learning ($\mathcal{L}_{cl}$) and OD prediction ($\mathcal{L}_{od}$) to the EMD loss proposed in CrowdDet:
\small
\begin{equation}
\label{eq:od}{}
\begin{aligned}
    \mathcal{L}(b_i)=\min_{\pi\in\Pi}\sum_{k=1}^{K}[\alpha\cdot(\mathcal{L}_{cls}(c_i^{(k)},g_{\pi_k})+\mathcal{L}_{reg}(I_i^{(k)},g_{\pi_k})) \\
    +\gamma\cdot\mathcal{L}_{cl}(c_i^{(k)},g_{\pi_k})+\eta\cdot\mathcal{L}_{od}({od}_i^{(k)},g_{\pi_k})],
\end{aligned}
\end{equation}
\normalsize
where $\alpha$, $\gamma$ and $\eta$ are coefficients of Eq.(9) in our paper.
Please refer to Eq.(3) in (Chu et al. 2020) for other details of the equation above.
During inference, we simply append the OD check to the original Set-NMS
to decide if a box should be suppressed when it is highly overlapped with anothor prediction derived from a different proposal.

\subsection{More discussions of the Consensus Learning}
In the main body of our paper, we ignore a small issue about the consensus learning (CL):
\emph{does the improvement come from additionally pasted objects in $\mathcal{B}^*_{ovl}$ ?}
To study this problem, we conduct ablation experiments in Table A~\ref{tab:cl-discuss}.
During training, we follow the same re-pasting process in the consensus learning pipeline but cancel the loss $\mathcal{L}_{cl}$ for pair-wise score alignment.
As shown by results in the third line, the extra pasting objects alone cannot improve the detector's performance,
which futher verifies the necessity of our design. 
%
%
\begin{table}[]
\small
\renewcommand{\tablename}{Table A}
\tabcolsep=0.27cm
\begin{tabular}{l|cc|cc}
\hline
\multirow{2}{*}{}           & MR$^{-2}$& AP@0.5                & MR$^{-2}$                  & AP@0.5                  \\ \cline{2-5}
                            & \multicolumn{2}{c|}{on \emph{Faster R-CNN}} & \multicolumn{2}{c}{on \emph{RetinaNet}}   \\ \hline
CCP                         & 41.11              & 87.75                 & 48.81                & 84.73                    \\
CCP+CL                      & 40.80              & 88.02                 & 47.93                & 84.85                   \\
CCP+CL w/o $\mathcal{L}_{cl}$ & 41.15              & 87.78               & 48.88                & 84.71                   \\ \hline
\end{tabular}
  \caption{
    \textbf{Futher discussions of the CL.}
    Experiments are conducted on Faster R-CNN and RetinaNet respectively.
    Results are reported on CrowdHuman val set.
  }
\label{tab:cl-discuss}
\end{table}

\subsection{Multi-Category Setting}
Since a typical crowded object detection task (and its corresponding open dataset) often includes only one category,
we make experiments on the pedestrians and cars respectively in the paper.
To further explore the effect of our method on the more general multi-category setting,
we train a Faster R-CNN on KITTI for ``pedestrian'' and ``car'' jointly, applying our CrowdAug.
As shown in Table A~\ref{tab:multi-cat}, improvements of detection performance are acquired on both categories.
%
%
\begin{table}[]
\small
\renewcommand{\tablename}{Table A}
\centering
\begin{tabular}{l|c|ccc}
\hline
{\small{\emph{on KITTI}}}                 {} &           & \small{Easy} & \small{Moderate} & \small{Hard}        \\\hline
{\multirow{2}{*}{\small{Faster R-CNN}}}   & ped   &  \small{97.30}  & \small{89.80}    & \small{79.41}        \\
{}              & car &  \small{87.85}  &  \small{76.31}        &  \small{70.23}             \\ \hline
{\multirow{2}{*}{\small{Faster R-CNN + \emph{ours}}}} & ped  &  \small{98.51}  & \small{91.26}    & \small{81.70}     \\
{}              & car  &  \small{88.61}    & \small{77.70}  & \small{72.11}        \\ \hline
\end{tabular}
\caption{\textbf{Experiments for the multi-category setting.}
          We use the categories of ``pedestrian'' and ``car'' in KITTI.
          AP@0.7 (\%) of \emph{easy}, \emph{moderate} and \emph{hard} objects are listed respectively.
        }
\label{tab:multi-cat}
\end{table}

\subsection{Additional Visualizations of OD}
\begin{figure*}[t]
\renewcommand{\figurename}{Figure A}
  \centering
  \includegraphics[width=1.\linewidth]{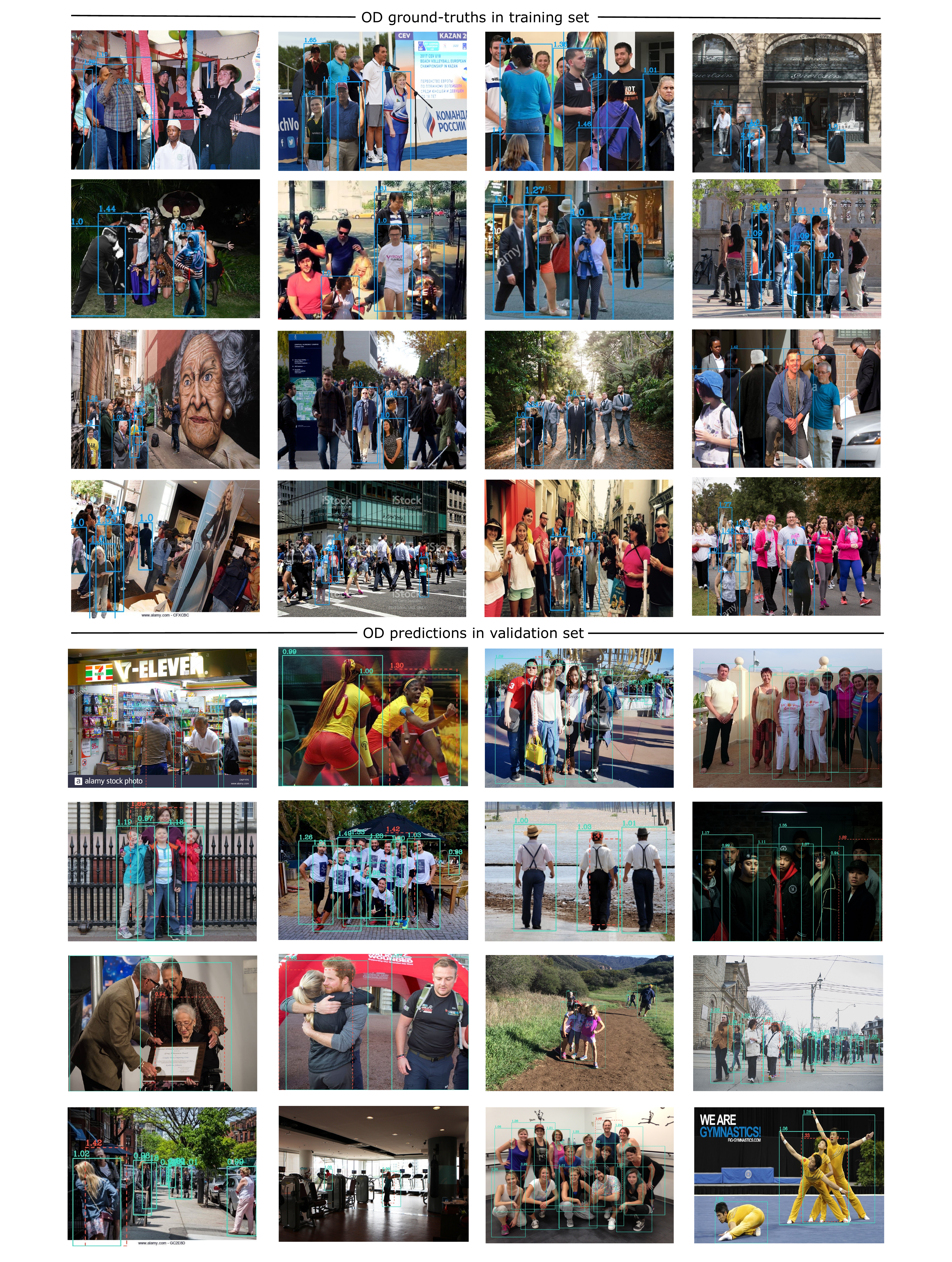}
  \vspace{-1.5\baselineskip}
  \caption{{\bf{Additional visualizations of OD.}}
            The value of overlay depth (OD) is marked at the top left corner of each box.
            The first four lines: OD ground truths generated by copy-paste process in training data (only objects pasted have the OD ground-truths).
            The last four lines: OD predictions of Faster R-CNN structure on the CrowdHuman val set.
            The \textit{red} dotted boxes denote the persons who are wrongly deleted by the original NMS while recalled by our OD-NMS.
            }
\label{fig:appendix-od-vis}
\end{figure*}

We make more comprehensive visualizations of the overlay depth (OD) in Fig. A\ref{fig:appendix-od-vis}.
In our method, learning of OD is in a semi-supervised manner.
During training, only the pasting objects (synthetic data) have OD ground-truths (the first four lines of Fig. A\ref{fig:appendix-od-vis})
while in inference, we expect the OD predictors perform well on natural data (the last four lines of Fig. A\ref{fig:appendix-od-vis}).
These visualizations suggest that our method can effectively learn the overlay depth and alleviate miss recall during de-duplication (red dotted boxes in Fig. A\ref{fig:appendix-od-vis}).
%
%
\subsection{Visualize the Detection Results}
\begin{figure*}[t]
\renewcommand{\figurename}{Figure A}
  \centering
  \includegraphics[width=1.\linewidth]{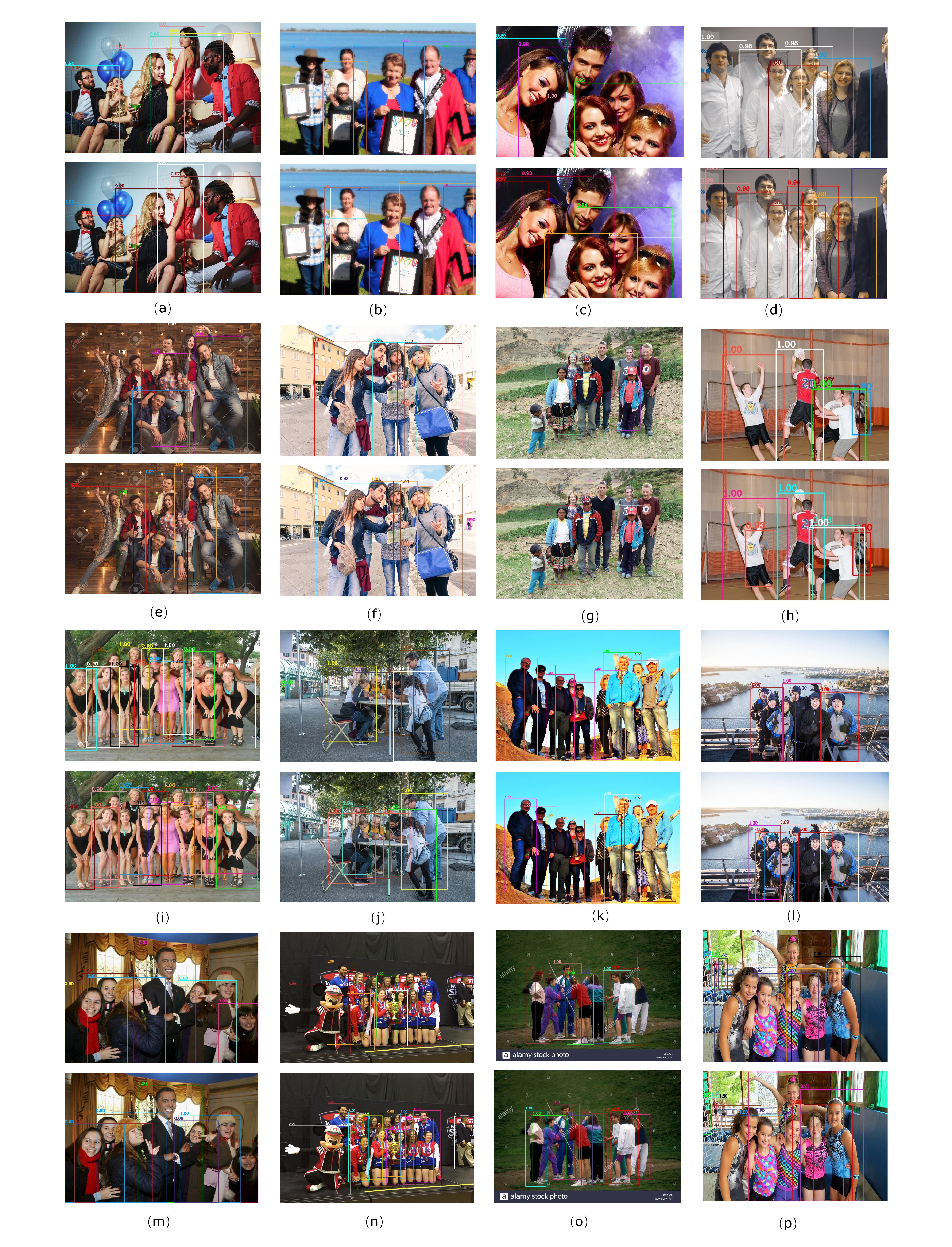}
  \vspace{-1.5\baselineskip}
  \caption{{\bf{Visualize the detection results.}}
            We visualize the detection results of the Faster R-CNN baseline (the top image for each pair)
            and those of our CrowdAug (the bottom image for each pair) on CrowdHuman val set.
            Confidence score is marked at the top left corner of each box.
            }
\label{fig:appendix-res-vis}
\end{figure*}

We visually compare the detection results of the baseline detector (Faster R-CNN) and our CrowdAug in Fig. A\ref{fig:appendix-res-vis}.
Qualitatively, we find two kinds of typical improvements. 
First, CrowdAug effectively avoids intermediate false boxes between objects (like (a), (b), (c), etc.), 
which we think is due to the better correlated confidence score with the IoU value.
Second, CrowdAug can reduce the miss recalls (like (d), (g), (h), etc.), 
which mainly comes from the more discriminative OD-NMS.
%
%
\makeatletter
\setlength{\@fptop}{0pt}
\makeatother

\end{document}